%% file: main.tex
\documentclass[journal]{IEEEtran}
\input{c_command}
\input{c_math_commands}

\input{c_dict}

\begin{document}
%
\title{ChiNet: Deep Recurrent Convolutional Learning for Multimodal Spacecraft Pose Estimation}
%
%
%


\author{Duarte~Rondao,
        Nabil~Aouf,
        and~Mark~A.~Richardson
\thanks{D. Rondao is a Postdoctoral Research Fellow with the Department of Electrical and Electronic Engineering at City, University of London, EC1V 0HB, UK (e-mail: \texttt{duarte.rondao@city.ac.uk}).}%
\thanks{N. Aouf is a Professor of Robotics and Autonomous Systems with the Department of Electrical and Electronic Engineering at City, University of London, EC1V 0HB, UK.}%
\thanks{M.A. Richardson is a Professor of Electronic Warfare with the Centre for Electronic Warfare, Information and Cyber at Cranfield University, Defence Academy of the United Kingdom, SN6 8LA, UK.}%
}

%
%

\markboth{SUBMITTED TO IEEE TRANSACTIONS ON AEROSPACE AND ELECTRONIC SYSTEMS}%
{Shell \MakeLowercase{\textit{et al.}}: Bare Demo of IEEEtran.cls for IEEE Journals}
%



\maketitle

\input{s_abstract}


\glsresetall

%
\IEEEpeerreviewmaketitle

\input{s_intro}

\input{s_related}
\input{s_method}
\input{s_experiments}
\input{s_conclusion}



\printbibliography

%
\begin{IEEEbiography}[{\includegraphics[width=1in,height=1.25in,clip,keepaspectratio]{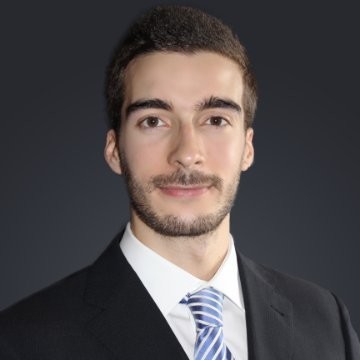}}]{Duarte Rondao}

is a postdoctoral research fellow in computer vision for space rendezvous in the Robotics and Machine Intelligence group at City, University of London. He has recently defended his PhD at Cranfield University on the topic ``Multispectral Navigation for Accurate Rendezvous Missions''. Despite being an early career researcher, Duarte has had almost 6 years of experience in the space sector, having worked on two different satellite missions: the European Student Earth Orbiter (ESEO) microsatellite, a joint initiative of ESA, AlmaSpace (now SITAEL), and the University of Bologna, Italy (ESEO was successfully launched into low Earth orbit in December 2018); and the ECOSat-III nanosatellite of the Centre for Aerospace Research at the University of Victoria, Canada, the successor to the group’s previous Canadian Satellite Design Challenge (CSDC) winning design.

\end{IEEEbiography}

\begin{IEEEbiography}[{\includegraphics[width=1in,height=1.25in,clip,keepaspectratio]{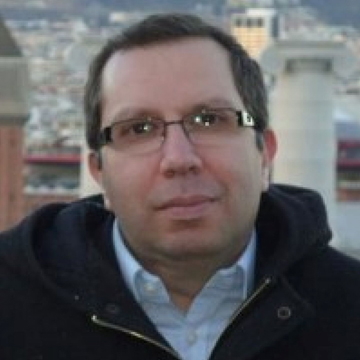}}]{Prof Nabil Aouf}

received his PhD from McGill
University in 2002 at the Electrical and Computer
Engineering Department. Currently, he is Professor
of Autonomous Systems and Machine Intelligence
at City University of London. He is the Director of
the Systems, Autonomy and Control (SAC) Centre
and the co-Director of the London Space Institute
(LSI) at City University of London. He also leads
the Robotics, Autonomy and Machine Intelligence
(RAMI) group and works very closely with industries that have a strong heritage in autonomous systems and space research. He has authored over 180 high calibre publications
in his domains of interest. His research interests are aerospace and defence
systems, information fusion and vision systems, guidance and navigation,
control, and autonomy of systems. He is an Associate Editor of 4 journals
including IEEE Transactions of Intelligent Vehicles.

\end{IEEEbiography}

\begin{IEEEbiography}[{\includegraphics[width=1in,height=1.25in,clip,keepaspectratio]{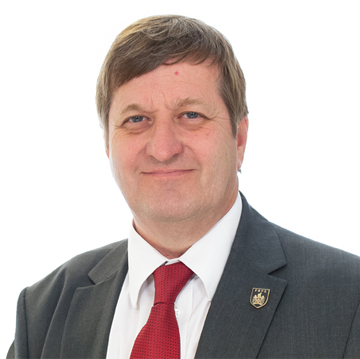}}]{Prof Mark A. Richardson}

has a BSc with First Class Honours in Physics from Imperial College London and is an Associate of the Royal College of Science. He has an MSc with Distinction in Applied Optics from Imperial College London, a Diploma of Imperial College, and PhD in Infrared Physics from Cranfield University. He has been at Cranfield University at the Defence Academy of the United Kingdom, Shrivenham, since 1989 and is currently the Pro-Vice-Chancellor of Cranfield Defence and Security. His research work is in the fields of Infrared Signature Simulation \& Modelling and EO\&IR Countermeasures. He has written over 300 classified and unclassified papers on these subjects, and holds a Classified Patent on a novel Infrared Camouflage Material. He is the editor and principal author of a book on battlefield surveillance technology and has acted as a consultant and defence analyst, on numerous occasions, to both the UK Ministry of Defence and commercial industry.

\end{IEEEbiography}




\end{document}

%% file: c_command.tex
\usepackage{siunitx}                
\usepackage{threeparttable}         
\usepackage{booktabs}               
\usepackage{adjustbox}              
\usepackage{breqn}                  
\usepackage{tabularx}
\usepackage[acronym]{glossaries-extra}  
\usepackage{glossary-mcols}             
\usepackage[en-GB]{datetime2}           
\usepackage{mathtools}                  
\usepackage{accents}                    
\usepackage[inline]{enumitem}           
\usepackage[
    bottom,
    hang,
    flushmargin]{footmisc}              

\usepackage{tikz}                   
\usepackage{graphicx}
\usepackage{subcaption}
\usepackage{transparent,import}     
\usepackage{placeins}               
\usepackage{epstopdf}
\usepackage[
    backend=biber,
    style=ieee,
    citestyle=numeric-comp,
    mincitenames=1,
    maxcitenames=1,
    maxbibnames=5,
    url=false,
    doi=false]
           {biblatex}
\usepackage{doi}                                    
\DeclareNameAlias{author}{family-given}
\DeclareNameAlias{editor}{family-given}
\DeclareNameAlias{translator}{family-given}
\newcommand*{\citet}{\textcite}
\newcommand*{\citep}{\parencite}

\newignoredglossary{hidden}
\glsdisablehyper
\newcolumntype{C}{>{\centering\arraybackslash}X} 
\AtBeginDocument{\sisetup{math-rm=\mathrm, detect-weight=true, binary-units = true, mode=text,range-phrase = --}}
\DeclareSIUnit\degree{deg}
\DeclareSIUnit\usd{USD}
\DeclareSIUnit\rpm{rpm}
\DeclareSIUnit\gauss{G}
\DeclareSIUnit\pixel{px}
\DeclareSIUnit\revolution{rev}
\newsavebox{\largestimage}
\DTMusemodule{british}{en-GB}
\hypersetup{
    colorlinks=true
}

\usepackage{lscape,pdflscape}
\usepackage{amsmath,amssymb,bm,amsfonts,amsthm,eucal}
\usepackage{multirow}
\usepackage{graphicx}

\usepackage{url}
\usepackage{makecell}

\usepackage{epsfig}

%% file: c_math_commands.tex
\newcommand{\figcustom}[1]{{\em (#1)}}

\def\Figref#1{Figure~\ref{#1}}
\def\figref#1{Fig.~\ref{#1}}
\def\Secref#1{Section~\ref{#1}}

\newcommand\citepos[1]{\citeauthor{#1}'s\ \parencite{#1}}

\def\acrconnect#1#2{%
\ifglsused{#1}
    {\glsxtrshort{#1}{#2}%
    }
    {\glsxtrshort{#1}{#2} (\glsxtrlong*{#1})\glsunset{#1}%
    }%
}
\def\acrcite#1#2{%
\ifglsused{#1}
    {\glsxtrshort{#1} \parencite{#2}%
    }
    {\glsxtrlong{#1} (\glsxtrshort*{#1} \cite{#2})\glsunset{#1}%
    }%
}
\def\acrplcite#1#2{%
\ifglsused{#1}
    {\glsxtrshortpl{#1} \parencite{#2}%
    }
    {\glsxtrlongpl{#1} (\glsxtrshortpl*{#1} \cite{#2})\glsunset{#1}%
    }%
}
\def\acrconnectcite#1#2#3{%
\ifglsused{#1}
    {\glsxtrshort{#1}{#2} \parencite{#3}%
    }
    {\glsxtrshort{#1}{#2} (\glsxtrlong*{#1} \parencite{#3})\glsunset{#1}%
    }%
}

\def\figref#1{Fig.~\ref{#1}}
\def\Figref#1{Figure \ref{#1}}
\def\eqref#1{Eq.~(\ref{#1})}
\def\Eqref#1{Equation (\ref{#1})}
\def\tabref#1{Tab.~\ref{#1}}
\def\Tabref#1{Table \ref{#1}}

\def\vone{{\bm{1}}}

\def\vtheta{{\bm{\theta}}}

\def\va{{\bm{a}}}

\def\vc{\protect{\bm{c}}}
\def\vd{{\bm{d}}}

\def\vf{{\bm{f}}}
\def\vg{{\bm{g}}}
\def\vh{{\bm{h}}}
\def\vi{{\bm{i}}}

\def\vo{{\bm{o}}}
\def\vp{\protect{\bm{p}}}
\def\vq{{\bm{q}}}
\def\vr{{\bm{r}}}

\def\vt{{\bm{t}}}

\def\vx{{\bm{x}}}

\def\vz{{\bm{z}}}

\def\eva{{a}}

\def\mK{{\bm{K}}}

\def\mR{{\bm{R}}}

\def\mT{{\bm{T}}}

\def\mW{{\bm{W}}}

\DeclareMathAlphabet{\mathsfit}{\encodingdefault}{\sfdefault}{m}{sl}
\SetMathAlphabet{\mathsfit}{bold}{\encodingdefault}{\sfdefault}{bx}{n}
\newcommand{\tens}[1]{\bm{\mathsfit{#1}}}

\def\tI{{\tens{I}}}

\def\sP{{\mathbb{P}}}

\def\sR{{\mathbb{R}}}

\def\sY{{\mathbb{Y}}}
\def\sZ{{\mathbb{Z}}}

\def\urr{{\textnormal{r}}}

\def\gL{{\mathcal{L}}}

\DeclareRobustCommand\undervec[1]{\underaccent{\vec}{#1}}

\def\Fc{\undervec{\bm{\mathcal{F}}}_{c}}

\def\Fo{\undervec{\bm{\mathcal{F}}}_{o}}
\def\Ft{\undervec{\bm{\mathcal{F}}}_{t}}

\newcommand{\normltwo}{L^2}

\newcommand{\sigmoidd}{\mathrm{sigm}}
\newcommand{\Sethree}[0]{\mathrm{SE(3)}}
\newcommand{\Sothree}[0]{\mathrm{SO(3)}}
\newcommand{\Sphere}[0]{\mathrm{S}}
\DeclareMathOperator*{\argmax}{arg\,max}
\newcommand{\given}[0]{\,\middle\vert\,}

\def\ct{\tau}
\def\dt{\kappa}

\ExplSyntaxOn

\NewDocumentCommand \irow { s o m }
 {
  \IfBooleanTF {#1}
   { \vectaux*{#3} }
   { \IfValueTF {#2} { \vectaux[#2]{#3} } { \vectaux{#3} } }
 }

\DeclarePairedDelimiterX \vectaux [1] {\lbrack} {\rbrack}
 { \, \dbacc_vect:n { #1 } \, }

\cs_new_protected:Npn \dbacc_vect:n #1
 {
  \seq_set_split:Nnn \l_tmpa_seq { , } { #1 }
  \seq_use:Nn \l_tmpa_seq { \enspace }
 }
\ExplSyntaxOff

%% file: c_dict.tex
\newacronym[category=hidden]{adr}{ADR}{active debris removal}
\newacronym[category=hidden]{asl}{ASL}{Autonomous Systems Laboratory}
\newacronym[category=hidden]{atv}{ATV}{Automated Transfer Vehicle}
\newacronym[category=hidden]{asmil}{ASMIL}{Autonomous Systems and Machine Intelligence Laboratory}

\newacronym[category=hidden]{bptt}{BPTT}{backpropagation through time}

\newacronym[category=hidden]{cad}{CAD}{computer-aided design}
\newacronym[category=hidden]{cnes}{CNES}{Centre National d'Études Spatiales, France}
\newacronym[category=hidden]{cnn}{CNN}{convolutional neural network}

\newacronym[category=hidden,longplural=Degrees-Of-Freedom]{dof}{DOF}{degree-of-freedom}
\newacronym[category=hidden]{dnn}{DNN}{deep neural network}
\newacronym[category=hidden]{drcnn}{DRCNN}{deep recurrent convolutional neural network}

\newacronym[category=hidden]{esa}{ESA}{European Space Agency}

\newacronym[category=hidden]{fc}{FC}{fully connected}
\newacronym[category=hidden]{fov}{FOV}{field of view}

\newacronym[category=hidden]{gnc}{GNC}{guidance, navigation, and control}
\newacronym[category=hidden]{gp}{GP}{guidance profile}
\newacronym[category=hidden]{gpu}{GPU}{graphics processing unit}

\newacronym[category=hidden]{ip}{IP}{image processing}

\newacronym[category=hidden]{lidar}{LIDAR}{LIght Detection and Ranging}
\newacronym[category=hidden,longplural=long short-term memories]{lstm}{LSTM}{long short-term memory}
\newacronym[category=hidden]{lvlh}{LVLH}{local-vertical-local-horizontal}
\newacronym[category=hidden]{lwir}{LWIR}{long wavelength infrared}

\newacronym[category=hidden]{ml}{ML}{machine learning}
\newacronym[category=hidden]{mse}{MSE}{mean squared error}
\newacronym[category=hidden]{mli}{MLI}{multi-layer insulation}

\newacronym[category=hidden]{nasa}{NASA}{National Aeronautics and Space Administration, USA}
\newacronym[category=hidden]{ncrv}{NCRV}{non-cooperative rendezvous}

\newacronym[category=hidden]{orb}{ORB}{Oriented FAST and Rotated BRIEF}

\newacronym[category=hidden]{pnp}{P$n$P}{perspective-$n$-point}

\newacronym[category=hidden]{ransac}{RANSAC}{random sample consensus}
\newacronym[category=hidden]{rv}{RV}{rendezvous}
\newacronym[category=hidden]{rgb}{RGB}{red-green-blue}
\newacronym[category=hidden]{rgbt}{RGBT}{red-green-blue-thermal}
\newacronym[category=hidden]{rnn}{RNN}{recurrent neural network}

\newacronym[category=hidden]{spec}{SPEC}{Satellite Pose Estimation Challenge}
\newacronym[category=hidden]{slam}{SLAM}{simultaneous localisation and mapping}
\newacronym[category=hidden]{speed}{SPEED}{Spacecraft PosE Estimation Dataset}
\newacronym[category=hidden]{sift}{SIFT}{Scale Invariant Feature Transform}
\newacronym[category=hidden]{surf}{SURF}{Speeded-Up Robust Features}

\newacronym[category=hidden]{tp}{TP}{tumbling profile}

\newacronym[category=hidden]{vo}{VO}{visual odometry}

%% file: s_abstract.tex
\begin{abstract}
This paper presents an innovative deep learning pipeline which estimates the relative pose of a spacecraft by incorporating the temporal information from a rendezvous sequence. It leverages the performance of \gls{lstm} units in modelling sequences of data for the processing of features extracted by a \gls{cnn} backbone. Three distinct training strategies, which follow a coarse-to-fine funnelled approach, are combined to facilitate feature learning and improve  end-to-end pose estimation by regression. The capability of \glspl{cnn} to autonomously ascertain feature representations from images is exploited to fuse thermal infrared data with \gls{rgb} inputs, thus mitigating the effects of artefacts from imaging space objects in the visible wavelength. Each contribution of the proposed framework, dubbed ChiNet, is demonstrated on a synthetic dataset, and the complete pipeline is validated on experimental data.
\end{abstract}

%% file: s_intro.tex
\section{Introduction}

\IEEEPARstart{S}{pacecraft} relative pose estimation is the problem of determining the rigid transformation between two
space bodies -- one of which is controllable and carries the navigation sensors -- in
terms of their relative position and attitude. This is a requirement for close-range \gls{rv} which has traditionally been solved using active sensors such as lidar \citep{fehse2003automated}; the task is significantly hampered when the target is said to be non-cooperative, i.e.\ it does not bear any supportive equipment towards the \gls{rv} \citep{wertz2003autonomous}.

\Gls{ncrv} operations involve the management of large relative velocities and minimal reaction times, justifying the need for autonomous operations and redundant sensors. As such, compact and lightweight passive digital cameras have become the cost-effective sensor for the task. Accordingly, the last couple of decades have focused on the development of robust \gls{ip} and \gls{ml} techniques to accurately estimate the target's six \gls{dof} pose from images obtained aboard the chaser \citep{cassinis2019review}. As the target is generally known beforehand, the followed strategies often choose to solve the model-to-image registration problem, under which the pose is retrieved via \acrcite{pnp}{szeliski2010computer} and \acrconnectcite{ransac}{-based}{fischler1981random} methods from correspondences between two-dimensional image features and three-dimensional model points. The challenge lies in robustly retrieving these correspondences in the face of hindering conditions such as shadows and sun glare, tumbling targets, or unknown initial poses. The former have been tackled in ground-based systems through multimodal sensing, but the fusion of each wavelength typically requires hand-crafted features, making its execution challenging \citep{mouats2015novel, beauvisage2020multimodal}.

\begin{figure}[t]
  	\centering
  	\begin{subfigure}[t]{\columnwidth}\centering
  		\includegraphics[trim=0 75 0 75,clip,angle=90,width=0.19\textwidth]{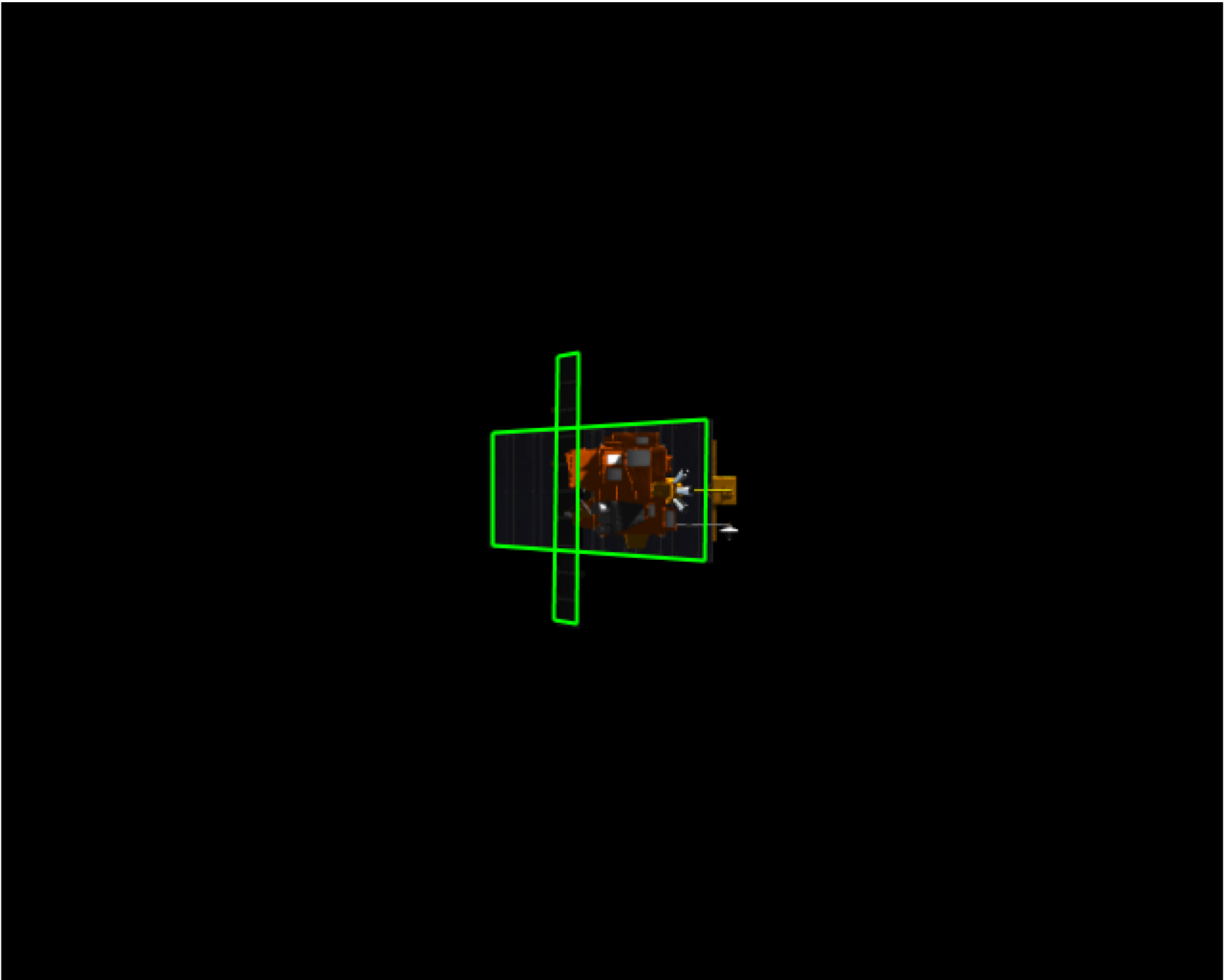}
  		\includegraphics[trim=0 75 0 75,clip,angle=90,width=0.19\textwidth]{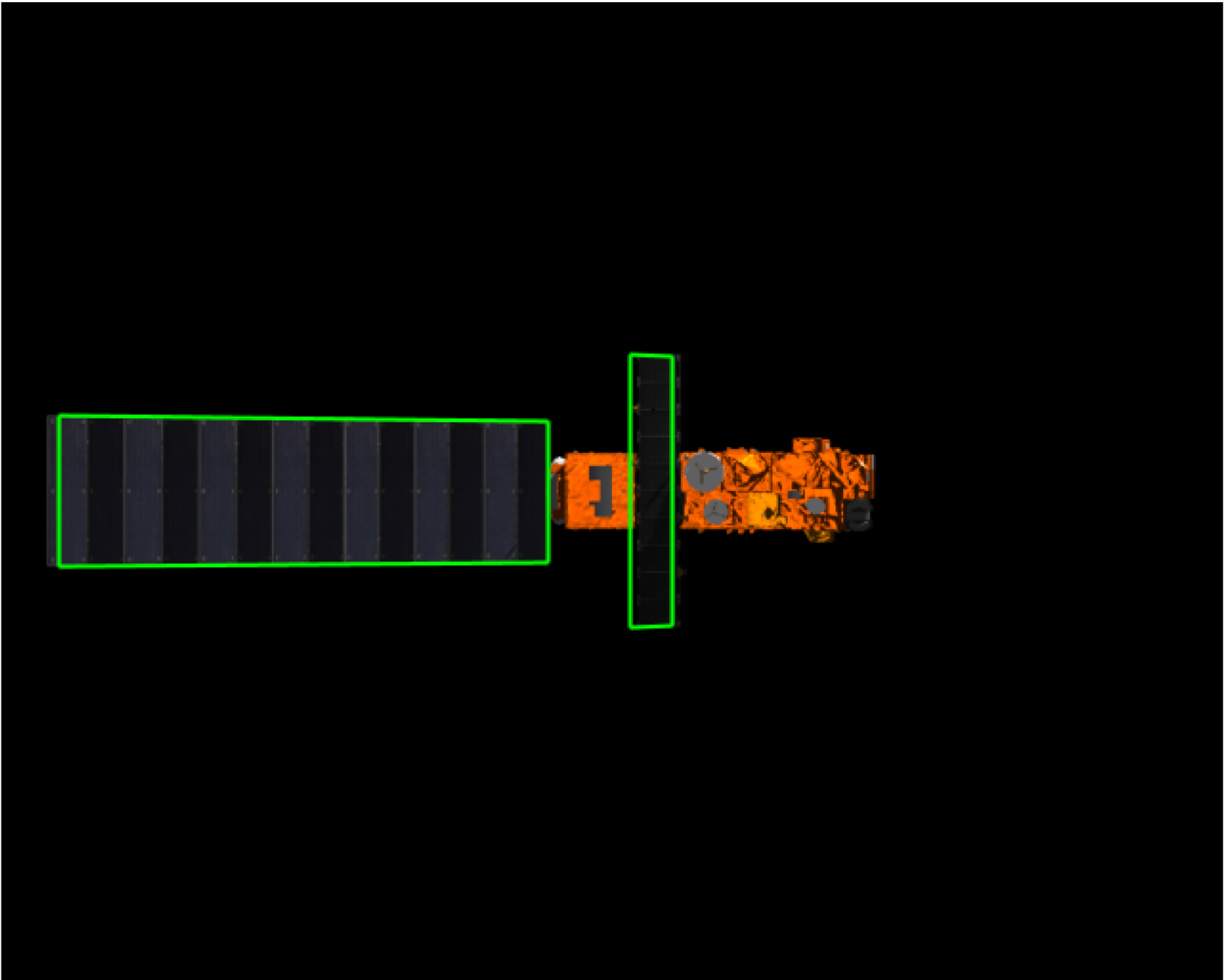}
  		\includegraphics[trim=0 75 0 75,clip,angle=90,width=0.19\textwidth]{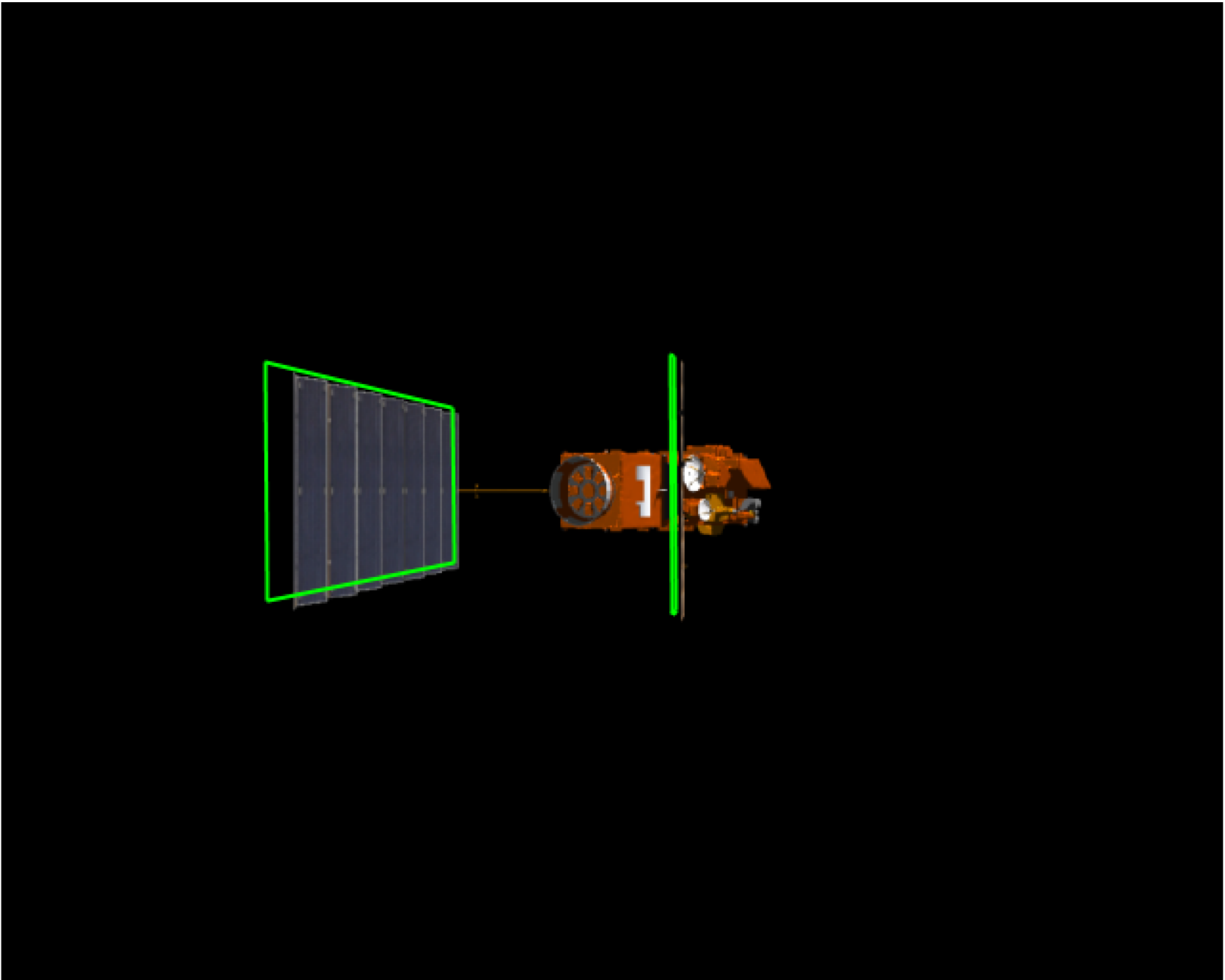}
  		\includegraphics[trim=0 75 0 75,clip,angle=90,width=0.19\textwidth]{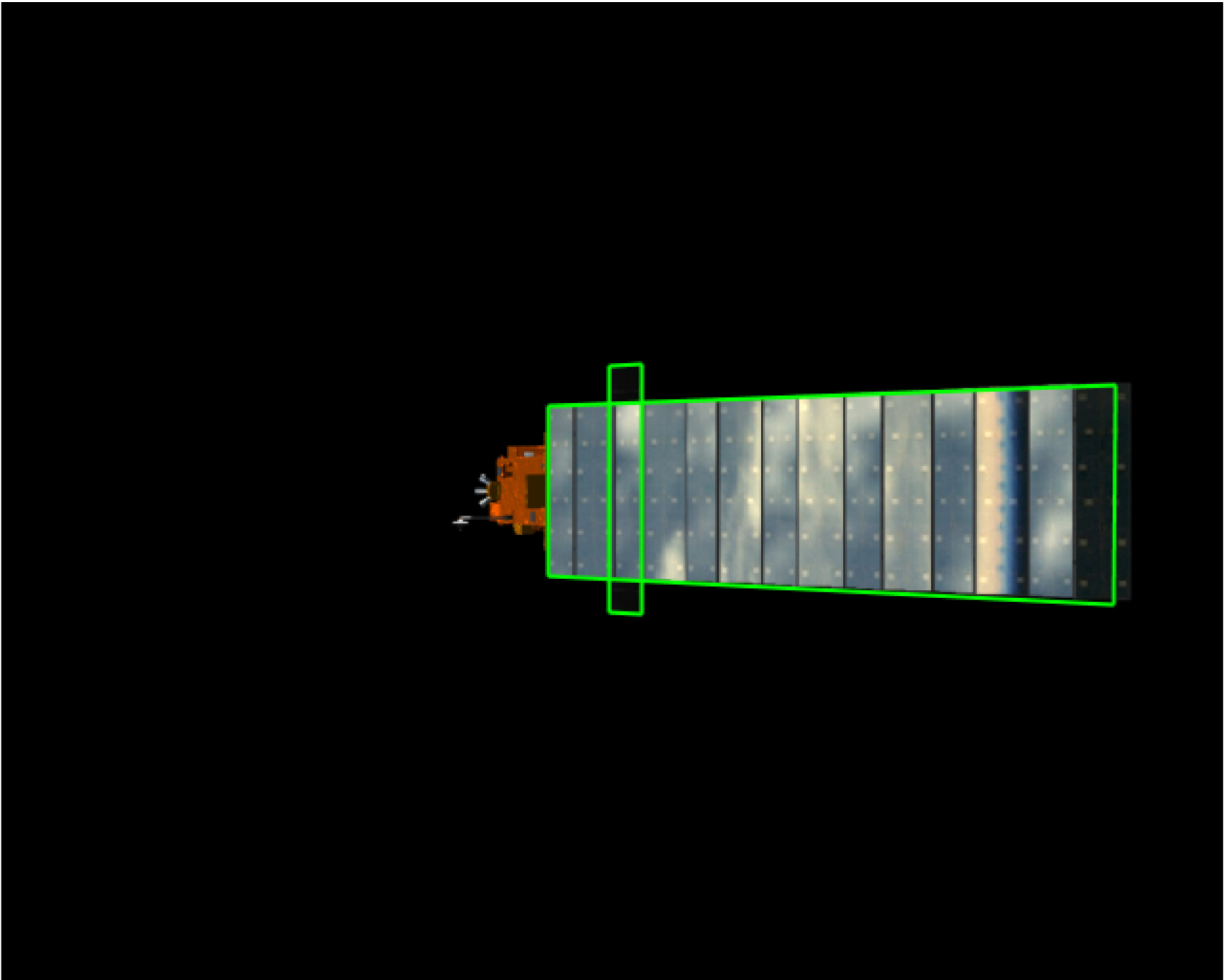}
  		\includegraphics[trim=0 75 0 75,clip,angle=90,width=0.19\textwidth]{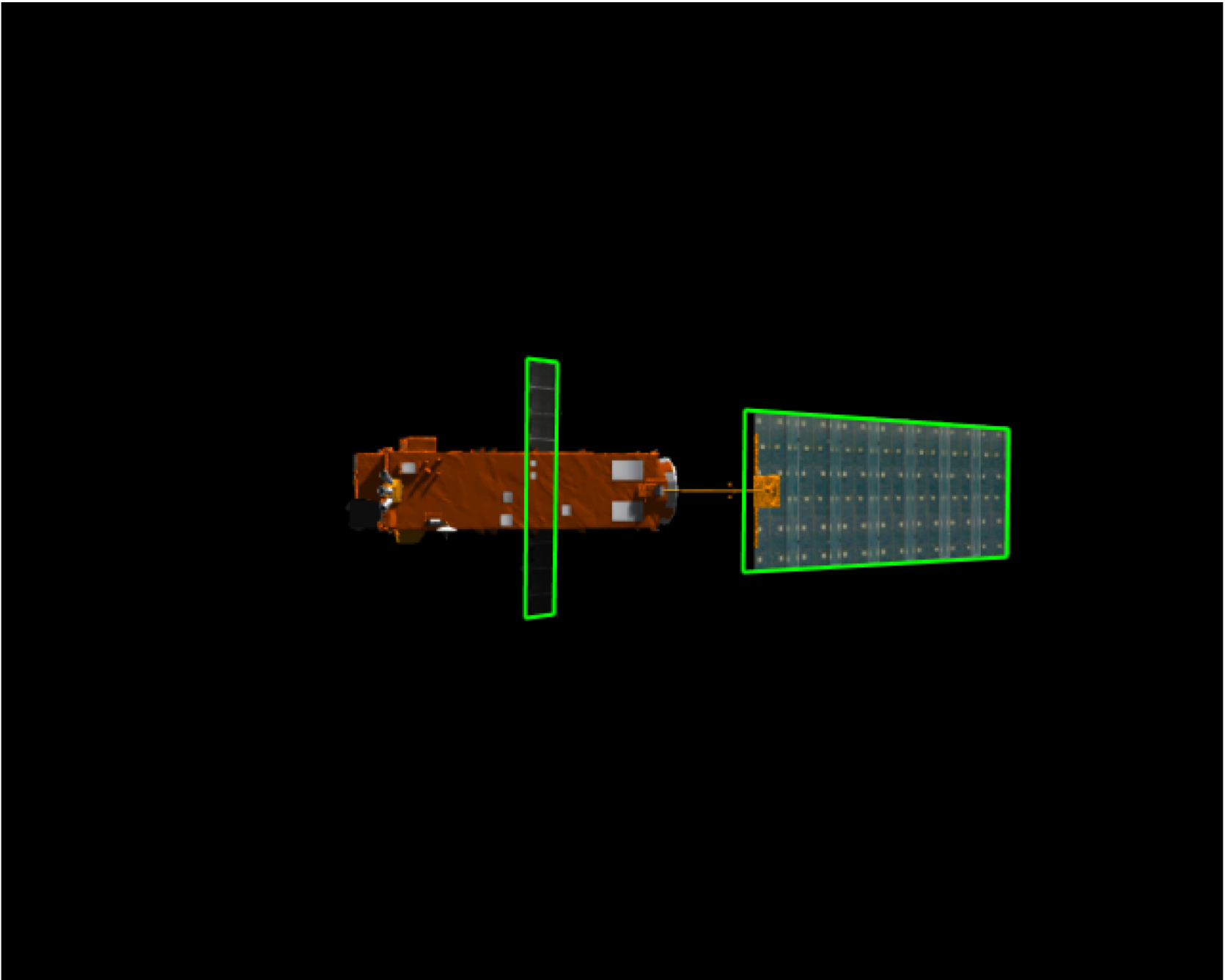}
      		\caption{\texttt{ASTOS/06}}
      		  \label{fig:intro-res-frames-test00-full-06}
  	\end{subfigure}\\ 
  	\begin{subfigure}[t]{\columnwidth}\centering
  		\includegraphics[trim=0 75 0 75,clip,angle=90,width=0.19\textwidth]{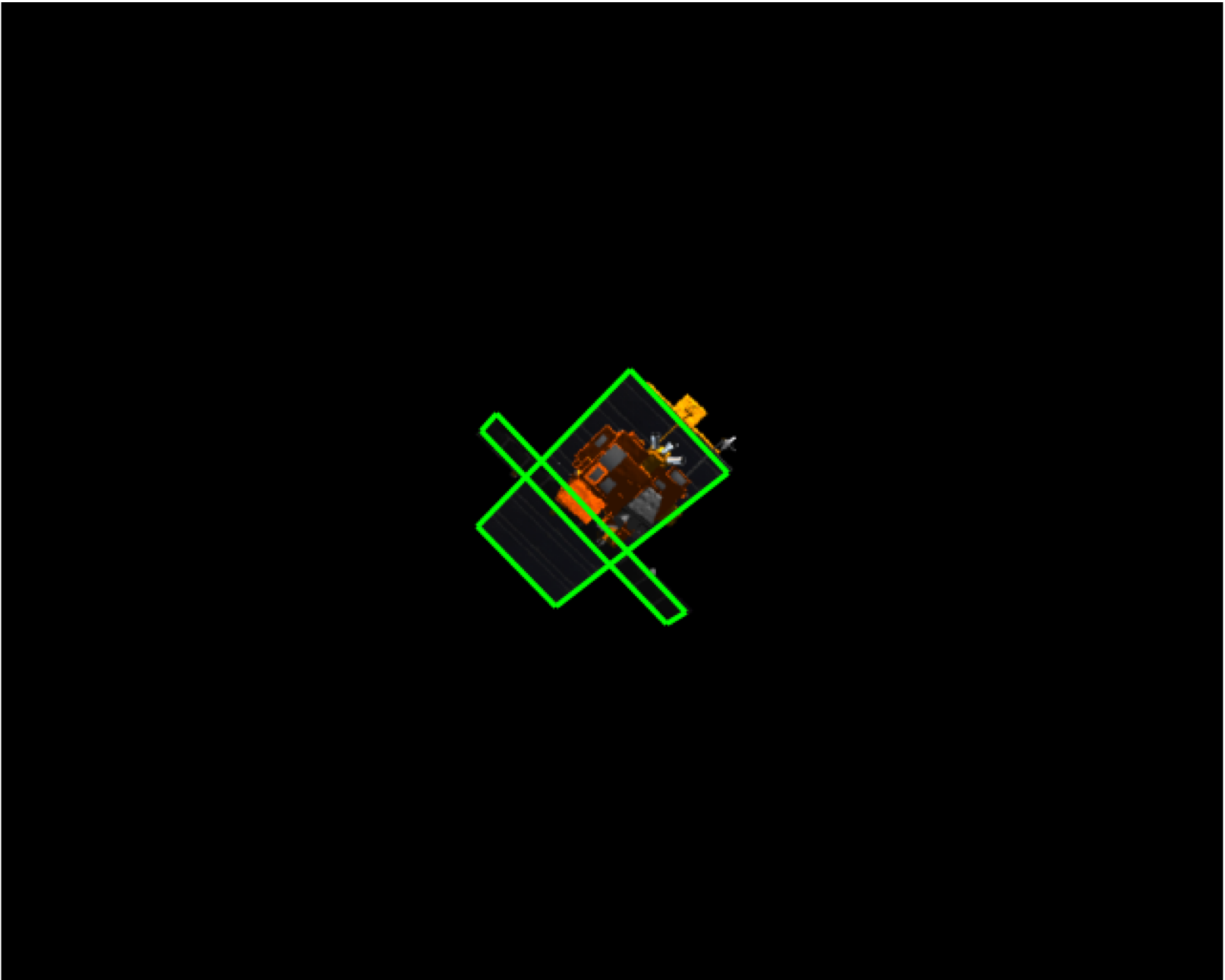}
  		\includegraphics[trim=0 75 0 75,clip,angle=90,width=0.19\textwidth]{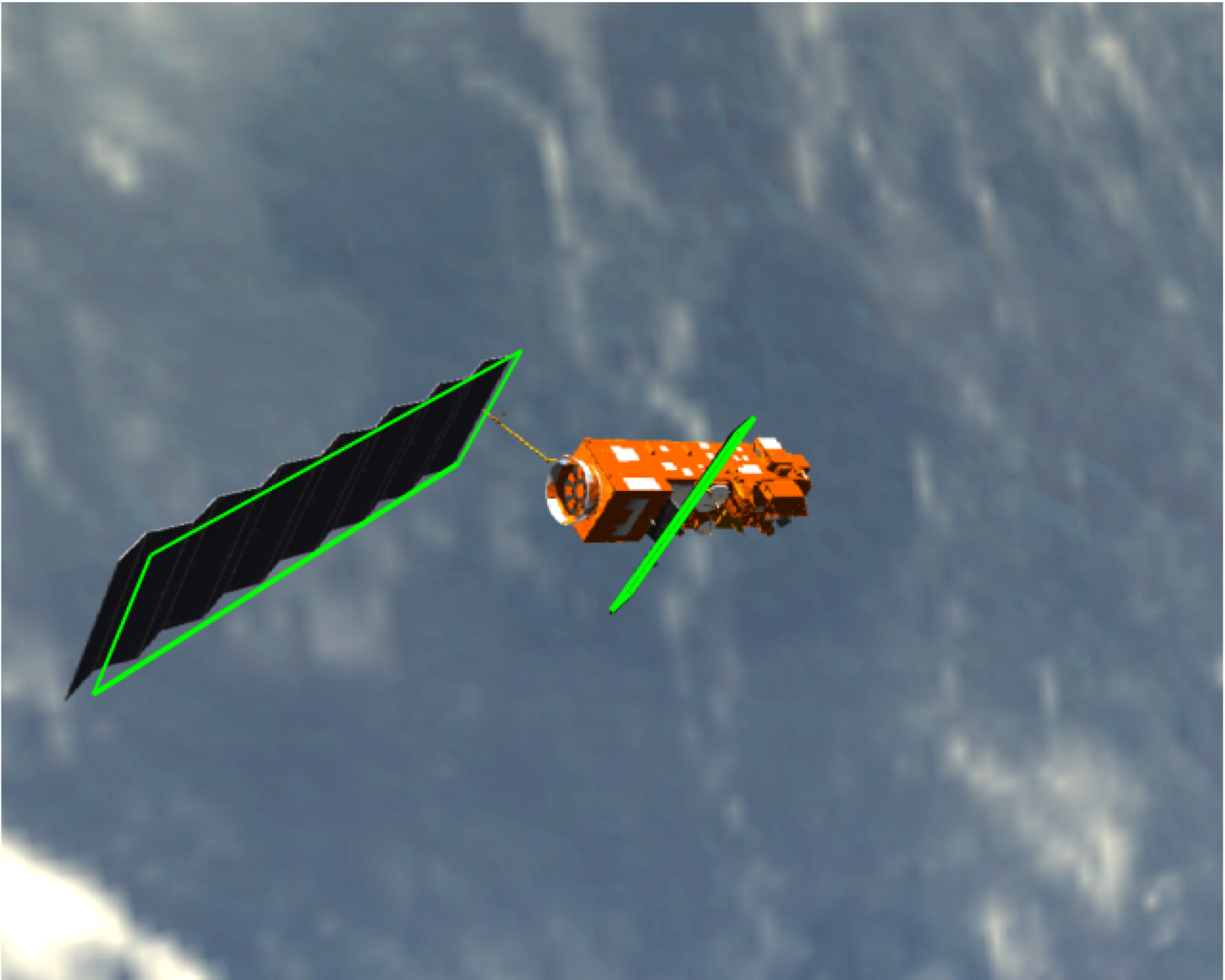}
  		\includegraphics[trim=0 75 0 75,clip,angle=90,width=0.19\textwidth]{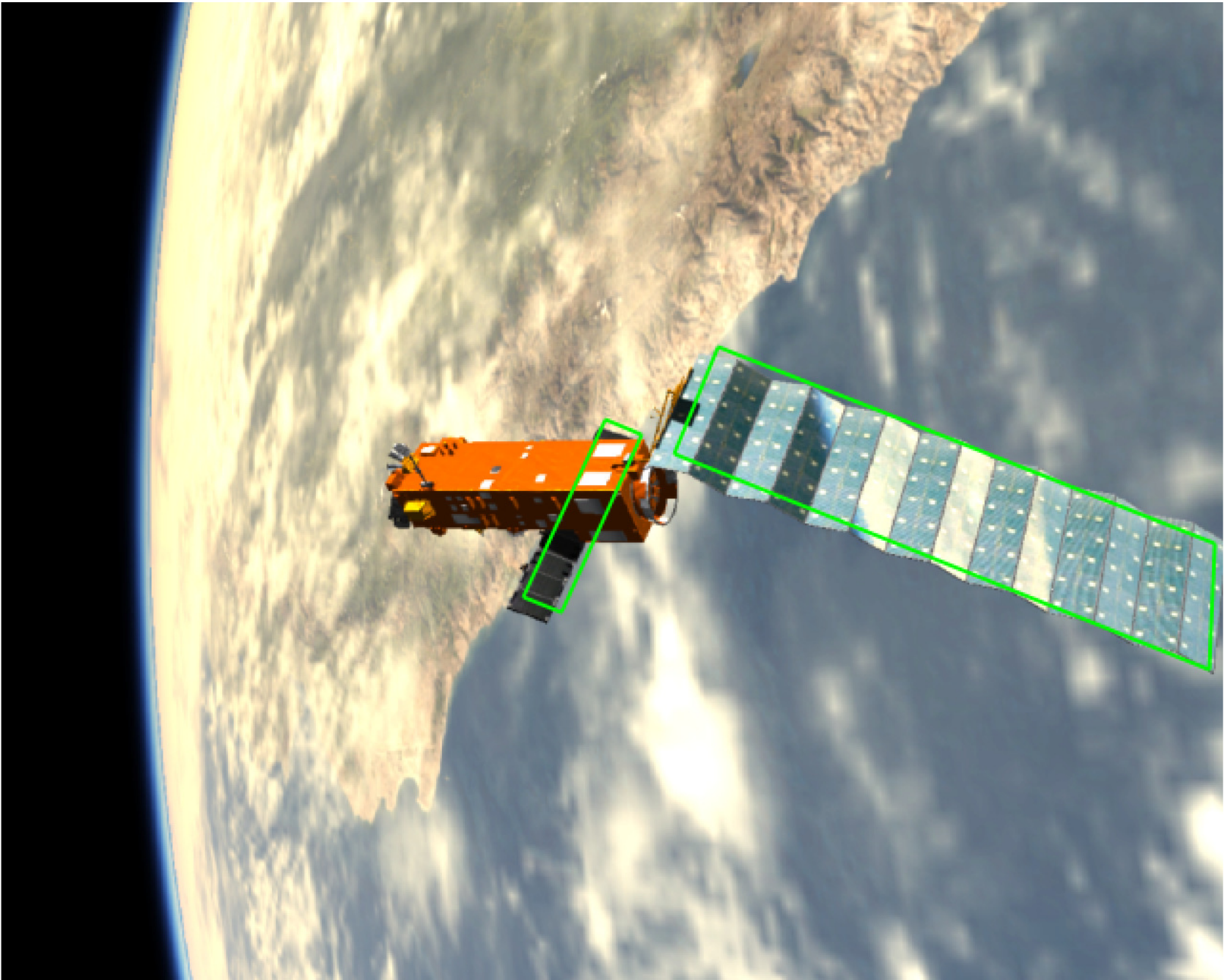}
  		\includegraphics[trim=0 75 0 75,clip,angle=90,width=0.19\textwidth]{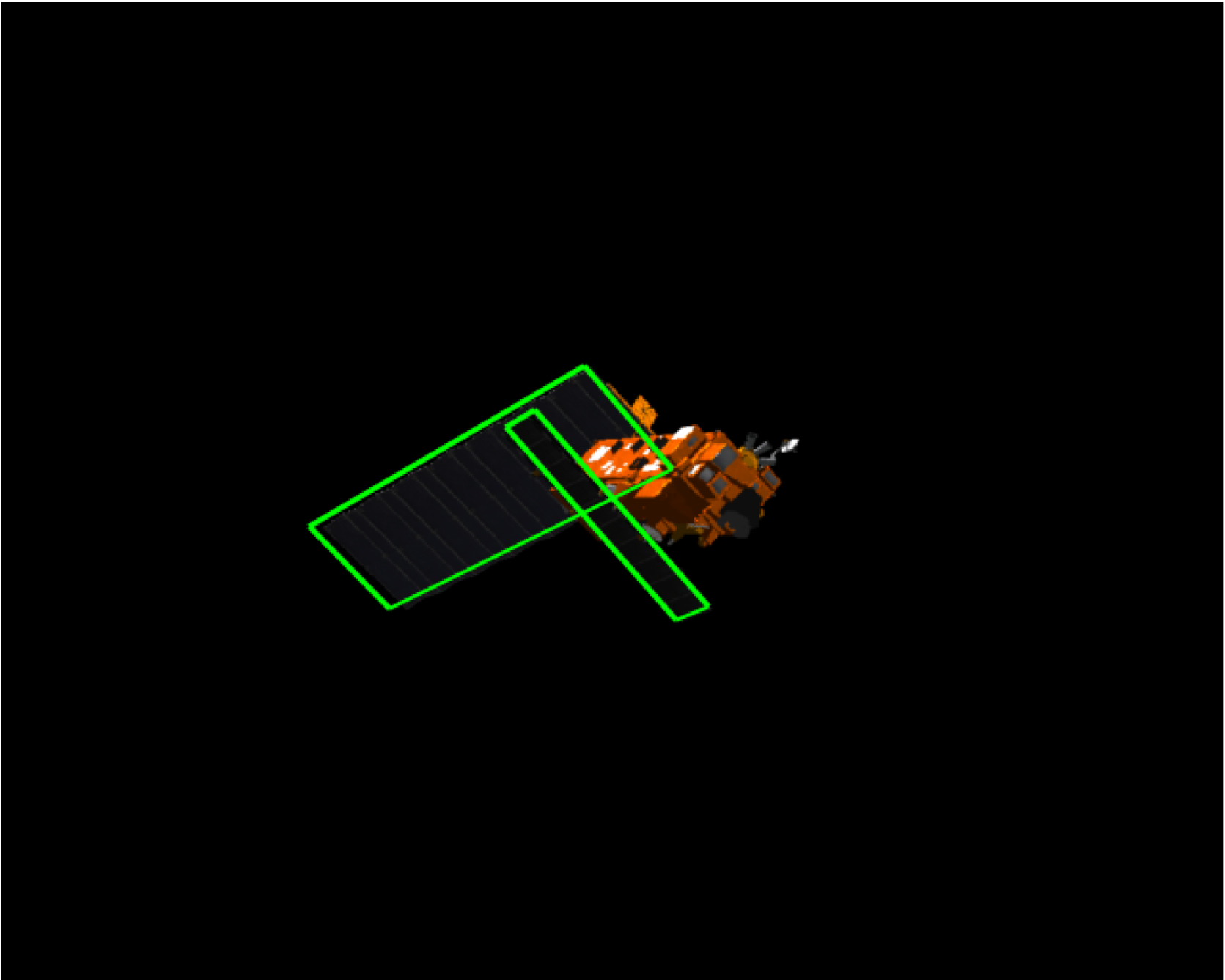}
  		\includegraphics[trim=0 75 0 75,clip,angle=90,width=0.19\textwidth]{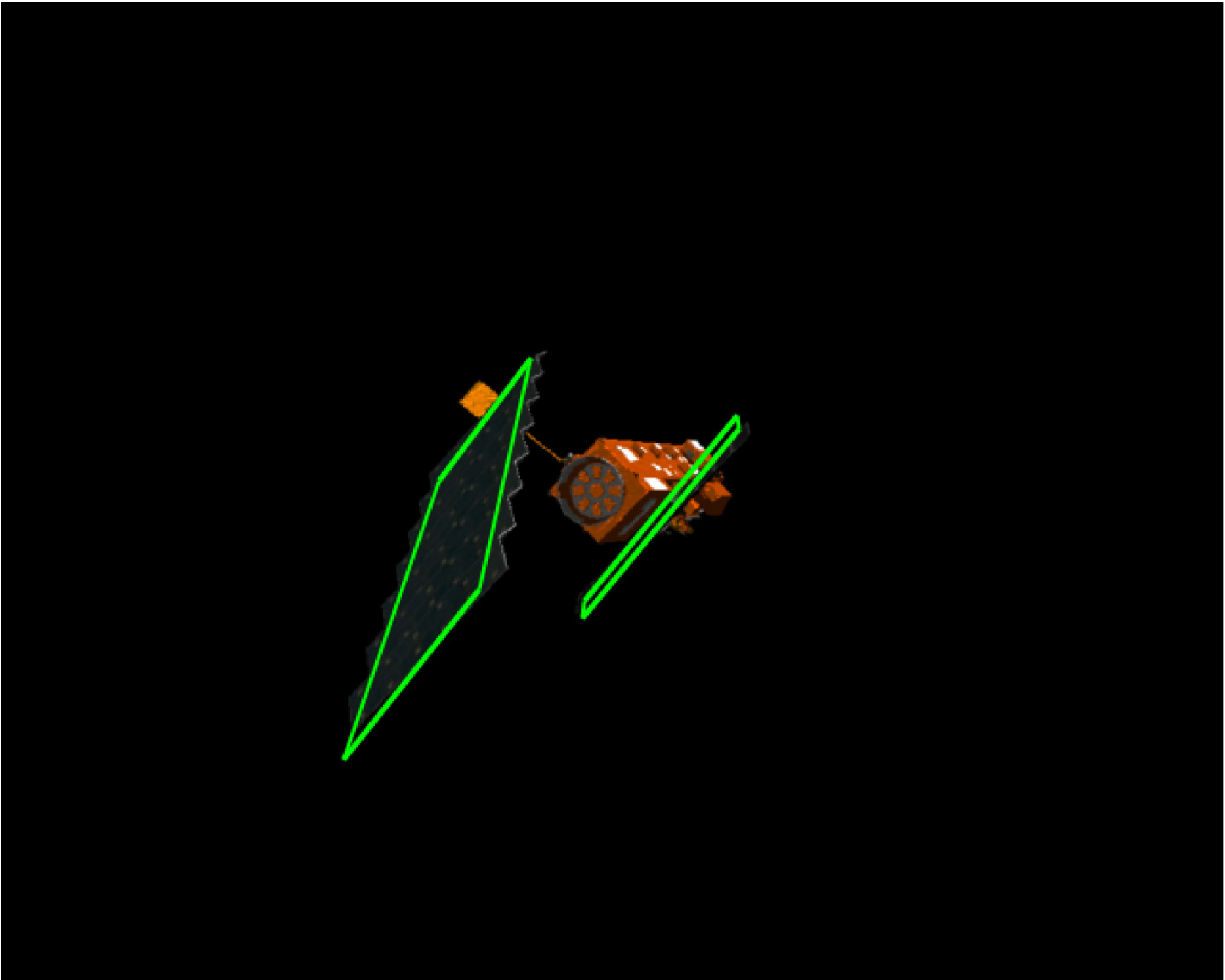}
  		\caption{\texttt{ASTOS/13}}
  		\label{fig:intro-res-frames-test00-full-13}
  	\end{subfigure}\\ 
  \caption{Qualitative results of the proposed method on two simulated \glsxtrshort{ncrv} sequences with Envisat from the Astos dataset. ChiNet provides continuous and robust pose estimation throughout the whole trajectories, explicitly taking into account information from previous frames.}
  \label{fig:intro-res-frames-test00-full}
\end{figure}

On the other hand, it represents an area with the potential of largely benefiting from \acrconnect{dnn}{-based} estimation methods. In particular, \acrplcite{cnn}{lecun1989handwritten} are naturally tailored to process such image inputs: here, the \gls{ip} task is shifted completely to the network, and the effort becomes concentrated towards parameter optimisation and data modelling, allowing for the generalisation of the model to a wider swath of imaging conditions. The popularity of \glspl{cnn} permeated onto the field of spacecraft relative pose estimation for rendezvous near the end of the past decade, mainly due to the \gls{esa} Kelvins \gls{spec},\footnote{\url{https://kelvins.esa.int/satellite-pose-estimation-challenge}.} where the vast majority (if not all) of the competitors used \acrconnect{dnn}{-based} approaches. \gls{spec} benchmarked the participating algorithms on the \acrcite{speed}{kisantal2020satellite}, which consists of images of the Tango satellite generated under unrelated randomised poses. However, during an \gls{rv} sequence, it is expected that the pose of the observed target continually varies as the operation progresses, i.e.\ the poses are correlated through time. 

This paper proposes the adoption of a \gls{rnn} module to process the features extracted by a \gls{cnn} front-end model and to exploit this temporal correlation between acquired image frames in \gls{ncrv} sequences. The resulting \gls{drcnn} architecture, dubbed ChiNet,\footnote{Pronounced ``kai-net'', the first term is an abbreviation of the Greek word ``chimera'', meaning ``something made up of parts of things that are different from each other''.} is shown to provide a smoother and lower-error estimate of the \num{6}-\gls{dof} pose when compared to a single \gls{cnn} (\figref{fig:intro-res-frames-test00-full} illustrates qualitative results on two \gls{ncrv} sequences). Furthermore, ChiNet proposes a new three-step training regimen to learn features in a coarse-to-fine manner, which is inspired from traditional \gls{ml} approaches. Lastly, ChiNet also explores the impact of multimodal sensing in the pose estimating by augmenting the number of input channels to the network with images from a \gls{lwir} camera, thus exploiting the natural ability of \glspl{cnn} to autonomously extract features from images. The following contributions are proposed, to the best of the authors' knowledge: 
\begin{enumerate*}
\item The work represents the first use of \glspl{rnn}, in particular \glspl{lstm}, to tackle the problem of spacecraft pose estimation for \gls{rv} using on-board cameras as the sole sensor;
\item It is also the first to explore the potential benefit of a multimodal sensor input for the task, in particular in the visible and \gls{lwir} modalities, leveraging the power of deep learning to formulate it as an optimal process and surpassing the hurdles of classical approaches; and
\item A multi-step optimisation approach to \gls{dnn} training is devised to facilitate the learning and reduce the overall estimation error.
\end{enumerate*}

The paper is organised as follows. \Secref{sec:relwork} surveys the literature to highlight pertinent related work. \Secref{sec:method} thoroughly details the methodology of each proposed contribution. \Secref{sec:exp} presents the attained results. Lastly, \Secref{sec:conc} shares the conclusions of the work.

%% file: s_related.tex
\section{Related Work}
\label{sec:relwork}

\noindent This section briefly summarises the existing model-based literature on spacecraft pose estimation with monocular cameras, i.e.\ when the target is known. It is broadly divided into two categories: methods based on geometry and methods based on learning (with a focus on \glspl{dnn}).

\subsection{Geometry-based Methods}

These methods estimate the \num{4x4} relative pose matrix $\mT = \mT_{ct}\in \Sethree$ relating the target body-fixed reference frame $\Ft$ to the camera frame $\Fc$, which is attached to the chaser, from model points $\vp^{(i)} \in \sR^3$ expressed in $\Ft$ and their image plane projections $\vz^{(i)} \in \sR^2$ expressed in $\Fc$, which are related according to the perspective projection model \citep{szeliski2010computer}:

\begin{equation}
    \vz^{(i)} = \pi\left(\mK 
    \begin{bmatrix}
        \mR & \vt
    \end{bmatrix}
    \begin{bmatrix}
    \vp^{(i)} \\ 1
    \end{bmatrix}
    \right), \label{eq:sec-related}
\end{equation}

\noindent where $i\in\{1,\ldots,N\}$ and the projective function ${ \pi(\va) \coloneqq \va_{1:2}/a_3 }$ has been defined. Here, $\mR,\vt$ are the \num{3x3} attitude matrix and \num{3x1} position vector composing $\mT$, and $\mK$ is the \num{3x3} intrinsic camera matrix accounting for the focal length $f$ obtained \textit{a priori} via calibration. 

\Eqref{eq:sec-related} can be solved in closed form for $N\geq 4$ using a \acrcite{pnp}{szeliski2010computer} solver, while using \acrcite{ransac}{fischler1981random} to reject spurious matches. Alternatively, it can be solved iteratively via robust estimation \citep{stewart1999robust}. Arguably, the biggest challenge resides in matching $\vz^{(i)}$ to $\vp^{(i)}$.

Tracking by recursion \citep{drummond2002realtime} was initially adopted as popular solution in which 3D control points from a \gls{cad} model of the target are projected onto the image using the expected pose accompanied by a gradient-based scan to locate the corresponding 2D feature. Initially limited to edge features \citep{kelsey2006vision}, the technique was later adapted to include other features such as colours \citep{petit2013robust} and keypoints \citep{petit2014combining} at the expense of requiring hardware acceleration to deal with complex models.

Conversely, tracking by detection entails an offline stage where a database of target feature points, whose positions on the surface are known, is built. Matching is then performed using heuristics exploiting the grouping of local model features and multiple hypotheses \citep{cropp2001pose, damico2014pose}; the pose and correspondence problems may also be solved concurrently at a higher computational cost \citep{shi2016spacecraft}. An alternative approach constructs a database by discretising the 3D object into 2D keyframes representing multiple viewpoints \citep{rondao2018multiview}, and then using local keypoint detectors and descriptors (e.g.\ \acrshort{sift} \citep{lowe2004distinctive}, \acrshort{surf} \citep{bay2006surf}, or the more modern \acrshort{orb} \citep{rublee2011orb}) to obtain the matches. 

Both tracking by detection and by recursion have been applied to spacecraft pose estimation in the \gls{lwir} \citep{shi2015uncooperative, gansmann20173d}, and to model-free estimation in general \citep{yilmaz2017using}. While the latter leverages the increased repeatability of \gls{lwir} features with respect to the visible band \citep{rondao2020benchmarking}, the former applications do not explicitly make use of such advantages, leaving a gap in the literature for this modality.

\subsection{Learning-based Methods}

These methods also estimate the pose $\mT$ but do not necessarily make use of \eqref{eq:sec-related} or local features, instead exploring patterns in training data to generalise towards previously unseen query images. A coarser estimation of $\mT$ can also be considered in order to initialise tracking by recursion methods or to reduce the search-space in tracking by detection.

Generally, global features (e.g.\ bags of keypoints, shapes, or even raw images) have been preferred for combination with a variety of \gls{ml} techniques ranging from nearest neighbour search \citep{comellini2021global} to unsupervised clustering \citep{petit20153d}, principal component analysis \citep{shi2017spacecraft}, Bayesian classification \citep{rondao2021robust}, and deep learning \citep{sharma2018pose}.

The recent prevalence of the latter with respect to the others originated from \gls{spec} in \num{2019}. As reported by \citet{kisantal2020satellite}, the majority of the participating teams used \glspl{cnn} to directly predict the relative pose of the target in an end-to-end, regressive fashion from each raw image (e.g.\ \citep{proenca2019deep}). The attitude estimation was noted to be the most challenging, and was improved in approaches which first included a target localisation step (e.g.\ \citep{sharma2020neural}). However, the best-performing entries, including those who won \num{1}\textsuperscript{st} and \num{2}\textsuperscript{nd} places, followed instead an indirect approach where the role of the \gls{cnn} was relayed completly towards the prediction of pre-selected keypoints in the image, which were then used with \gls{pnp} to recover the pose \citep{chen2019satellite}.

After \gls{spec}, published \acrconnect{dnn}{-based} work has seldom considered actual rendezvous trajectories \citep{cassinis2021evaluation}, continuing to focus instead on individual greyscale images of \gls{speed} \citep{harvard2020spacecraft,huo2020fast,piazza2021deep,garcia2021lspnet}. In either case, the proposed strategies consist in using a \gls{cnn} for keypoint detection for use with \gls{pnp}. Additionally, the contribution of modalities beyond the visible remains to be fully investigated \citep{hogan2021using}.

Contrary to the above examples, ground-based applications have recently adopted the use of \glspl{rnn} combined with features extracted by \gls{cnn} front-ends to model the intrinsic motion dynamics from sequences of imaging data rather than individual inputs \citep{clark2017vinet,wang2017deepvo}; more specifically, these proposed \acrconnectcite{lstm}{-based}{hochreiter1997long} \glspl{drcnn} for \gls{vo} to estimate a car's egomotion. \citet{stamatis2020deeplo} introduced DeepLO, which followed the same philosophy for lidar-based relative navigation with a non-cooperative space target. Lidar data was preprocessed by quantisation and projection onto each plane in the target body frame of reference, thus creating three 2D depth images to be processed by a regular \gls{cnn}. 

%% file: s_method.tex
\section{Methodology}
\label{sec:method}

\noindent This section describes in detail the proposed \gls{drcnn} framework for end-to-end spacecraft pose estimation. The \gls{cnn} and \gls{rnn} modules are both described, as well as the multistage optimisation strategy to train them.

\subsection{System Architecture}

\begin{figure*}[t]
	\centering
	\includegraphics[width=\textwidth]{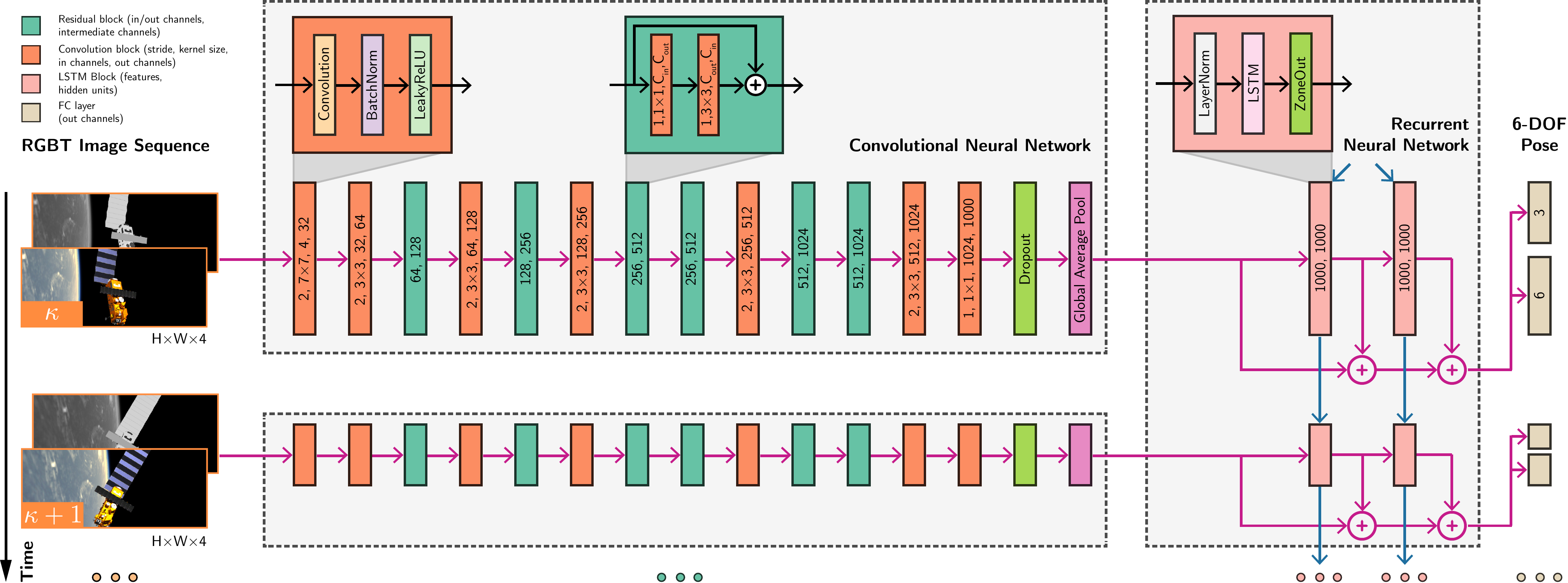}
 \caption{ChiNet system overview. The proposed \glsxtrshort{drcnn} architecture performs end-to-end spacecraft pose estimation from a sequence of multimodal \glsxtrshort{rgbt} image inputs of arbitrary size.}
\label{fig:method-chischeme}
\end{figure*}

The results from \gls{spec} have shown promising results in the use of \glspl{cnn} for the task. However, the current literature treats each incoming image as a separate input, thus ignoring the intrinsic temporal correlation between them. Therefore, the main focus here is the investigation of the feasibility of a \gls{drcnn} for estimating the pose in rendezvous sequences. The problem has been previously studied by \citet{stamatis2020deeplo} for \gls{vo} with lidar map inputs, but not for images. Furthermore, \gls{vo} is concerned with estimating the motion between two time-consecutive images, but during an \gls{rv} a single acquired image contains enough information relating $\Ft$ to $\Fc$. This work recognises this as a requirement and as such considers it for the \gls{drcnn} formulation.

The architecture of the proposed framework is schematically depicted in \Figref{fig:method-chischeme}. The pipeline takes a four-dimensional \gls{rgbt} image formed from the channel-wise concatenation of a visible image and a \gls{lwir} image. This multimodal image is then processed by a \gls{cnn}, whose learned output features are modelled temporally (along the vertical axis in the figure) with an \gls{rnn}. Two \gls{fc} layers convert the output into position and attitude values forming the 6-\gls{dof} pose. Note that, unlike in DeepVO or DeepLO, ChiNet receives only a single target snapshot at a time, thus predicting the complete relative pose for each time-step $\tau=\tau_\kappa$. Additionally, since the front-end is fully convolutional, the network is capable of receiving inputs of arbitrary spatial dimensions (i.e.\ any width and height).

\subsection{Deep Feature Extraction with Convolutions}

\gls{cnn} front-ends for feature extraction are typically chosen to be large but powerful architectures, such as ResNet \citep{he2016deep} or Inception-v3 \citep{szegedy2016rethinking}, and \acrconnect{spec}{-admitted} architectures were no exception. On the other hand, these networks are also characterised by elevated processing times and are potentially prone to overfitting due to their high number of parameters.

To mitigate this, ChiNet adopts the Darknet-19 architecture (backbone of the YOLO object detector \citep{redmon2017yolo9000}), with some modifications (\figref{fig:method-chischeme}, centre). First, the \num{3x3} kernel size on the first convolutional layer is replaced by a \num{7x7} one to adapt to image inputs larger than \SI{224 x 224}{\pixel}. Second the network is modernised (bringing it closer to Darknet-53 \citep{redmon2018yolov3}) by replacing all max pooling layers with a stride of \num{2} in the preceding convolution. Whereas the former is a fixed operation, the latter is learned, which further contributes to the adaptability of the network to the task at hand. In addition, residual connections are introduced but only in the channel expansion-contraction layers (green blocks in \figref{fig:method-chischeme}), thus avoiding the need to add extra \num{1x1} convolutions to keep the dimensions consistent. Lastly, a dropout layer \citep{hinton2012improving} with probability ${p = 0.2}$ is added to further prevent overfitting.

\subsubsection{Optimal Low-Level Sensor Fusion}

ChiNet preprocesses images acquired separately by each camera via concatenation along the channel dimension, forming a four-channel \gls{rgbt} image which the network takes as input. The first convolutional layer entails a weighted sum of the pixels in each channel, outputting new activation maps that effectively encompass the fused information. This is equivalent to a pixel (or low-level) fusion of the inputs resulting in a series of multimodal images upon which feature extraction is to be performed. Furthermore, these weights are not predefined but learned in the context of the network training procedure, thus being optimal in the sense of minimising the objective loss. This philosophy has been previously explored in \gls{vo} applications using traditional \gls{ip} techniques such as intensity level thresholding and discrete wavelet transforms, showing promising results \citep{poujol2015visiblethermal}. ChiNet's approach, however, bypasses the need of manually developing a potentially sub-par weighing strategy to combine the multiple input modalities.

\subsection{Temporal Sequence Modelling with \glsfmtshortpl{lstm}}

The features learned by the \gls{cnn} are post-processed by a deep \gls{rnn} module that models the intrinsic temporal correlations coming from an ordered sequence of image inputs. This addition is expected to be beneficial to the problem of spacecraft pose estimation due to the inherent relative motion dynamics entailed, and the estimate of the solution for the current frame can benefit from the knowledge of previous frames: even more than in ground-based applications, the perceived motion of a space target during \gls{rv} is not likely to change abruptly but is a smooth function of the previous states.

\begin{figure}[t]
	\centering
	\includegraphics[width=\columnwidth]{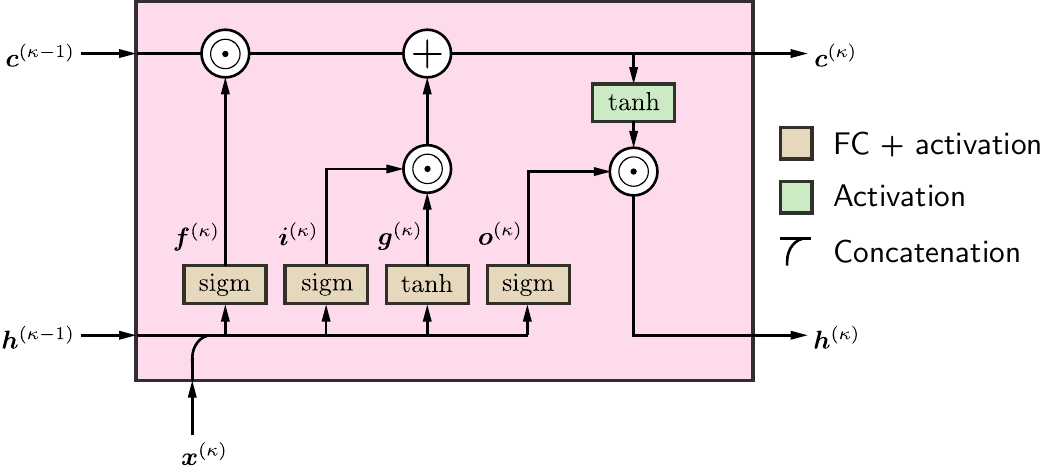}
 \caption{Block diagram of a \glsxtrshort*{lstm} recurrent memory unit. $\mathrm{sigm}$ and $\mathrm{tanh}$ denote the sigmoid and hyperbolic tangent activation functions, respectively; $\odot$ and $+$ denote element-wise product and addition, respectively.}
\label{fig:method-lstm}
\end{figure}

ChiNet's recurrent feature post-processing module is based on the \gls{lstm} architecture \citep{hochreiter1997long}. \glspl{lstm} were designed in an attempt to combat vital flaws in the capability of vanilla recurrent cells to model long sequences, as they suffered from vanishing and exploding gradients. The \gls{lstm}'s ability to learn long-term dependencies is owed to its gated design that determines which sectors of the previous hidden state should be kept or discarded in the current iteration. This is achieved not only in combination with the current input, processed by four different units, but also by a cell state which acts as an ``information motorway'' that bypasses the cells. The \gls{lstm} structure is illustrated in \Figref{fig:method-lstm}.

The design of the \gls{rnn} is schematically depicted in \Figref{fig:method-chischeme} (right). The \gls{cnn} features are fed to two stacked \gls{lstm} layers with \num{1000} hidden states each; stacked \gls{lstm} layers have been previously adopted for architectures such as DeepVO \citep{wang2017deepvo} and DeepLO \citep{stamatis2020deeplo} and shown empirically to help in modelling complex motion dynamics.

Unlike \gls{fc} or convolutional layers, data normalisation in \glspl{lstm} must be done internally due to the gated system topology. Batchnorm would be impractical both in terms of time and memory consumption since since this would require fitting one layer per time-step and storing the statistics of each one during training. In opposition, layer normalisation \citep{ba2016layer} is instead employed by computing the mean and variance across all the features of the $i$-th layer rather than across the batch dimension.

A second nuanced aspect pertains to dropout, typically applied as a binary mask to randomly nullify some of a layer's activations. In the case of \glspl{lstm}, however, stochasticity should be applied in the recurrent loop. More than that: rather than following a potentially naive dropout philosophy, ChiNet employs zoneout \citep{krueger2017zoneout}, which was specifically designed for \glspl{rnn}. In zoneout, the values of the hidden state $\vh^{(\dt)}$ and memory cell $\vc^{(\dt)}$ are randomly expected to either maintain their previous value or are updated in the usual manner. The modified \gls{lstm} equations thus become:

\begin{dmath}
\begin{bmatrix}
\tilde{\vf}^{(\dt)}\\
\tilde{\vi}^{(\dt)}\\
\tilde{\vo}^{(\dt)}\\
\tilde{\vg}^{(\dt)}
\end{bmatrix} = \mathrm{LN}\left(\mW^h \vh^{(\dt-1)};\gamma_1,\beta_1\right) + \mathrm{LN}\left(\mW^x \vx^{(\dt)};\gamma_2,\beta_2\right),
\end{dmath}
\begin{dmath}
\vc^{(\dt)} = \vd^{c,(\dt)} \odot \vc^{(\dt-1)} + \left(\vone - \vd^{c,(\dt)}\right) \odot \left(\vf^{(\dt)} \odot \vc^{(\dt-1)} + \vi^{(\dt)} \odot \vg^{(\dt)}\right),
\end{dmath}
\begin{dmath}
\vh^{(\dt)} = \vd^{h,(\dt)} \odot \vh^{(\dt-1)} + \left(\vone - \vd^{h,(\dt)}\right) \odot \left(\vo^{(\dt)} \odot \tanh\left\{\mathrm{LN}\left[\vc^{(\dt)};\gamma_3,\beta_3\right]\right\}\right),
\end{dmath}

\noindent where $\vf,\vi,\vo,\vg$ are the forget, input, output, and modulation gates, respectively; $\vh$ is the hidden state; $\vx$ is the input; $\mW^{h^\top} = \irow{\mW^{hf^\top},\mW^{hi^\top},\mW^{ho^\top},\mW^{hg^\top}}$ is the recurrent weights matrix; $\mW^{x^\top} = \irow{\mW^{xf^\top},\mW^{xi^\top},\mW^{xo^\top},\mW^{xg^\top}}$ is the input weights matrix; $\vf = \sigmoidd(\tilde{\vf})$; $\vi = \sigmoidd(\tilde{\vi})$; $\vg = \tanh(\tilde{\vg})$; $\vo = \sigmoidd(\tilde{\vo})$; $\sigmoidd$ is the sigmoid nonlinear activation function; $\tanh$ is the hyperbolic tangent activation function; $\mathrm{LN}$ denotes layer normalisation with scale $\gamma$ and offset $\beta$; $\vd^{c}, \vd^{h}$ are the binary cell and hidden state zoneout masks, respectively; $\vone$ is a vector of ones of appropriate length; the superscript $(\dt)$ denotes a variable at time-step $\ct=\ct_\dt$; and $\odot$ denotes an element-wise product operation.

Residual connections have also been implemented (see \figref{fig:method-chischeme}, right), drawing inspiration from the \gls{cnn} front-end itself. During preliminary experiments, it was found that the addition of residual connections to the \glspl{lstm} in ChiNet resulted in faster training convergence and overall lower pose estimation error.

\subsection{Multistage Optimisation}

Instead of pursuing an indirect approach (i.e.\ \gls{dnn} to predict keypoints followed by \gls{pnp}), ChiNet provides an end-to-end, direct method to retrieve the pose. The former has been shown to produce the lowest error estimates in \gls{spec}, suggesting that the latter may be harder to train. To mitigate this and lower the overall error in end-to-end approaches, a multistage, coarse-to-fine approach is proposed and described in this section.

\subsection*{Stage 1}

The objective of Stage 1 is to emulate the benefits of transfer learning \citep{goodfellow2016deep}, in which the network is pre-trained on a set of tasks involving a large dataset and then used to initialise a same-sized network to solve the purported task that generally has fewer training examples. Transfer learning is advantageous for \glspl{cnn} as these normally entail millions of parameters and thus may converge towards a suboptimal solution if the training data is not diverse enough. 

A subset of \num{1000} object categories of ImageNet \citep{deng2009imagenet} is the typical go-to choice for pre-trained networks. However, the data is composed of \gls{rgb} images and thus cannot be expanded for use with multimodal data. As such, a strategy to pre-train a \gls{cnn} by artificially augmenting the number of samples based only on the actual training dataset is proposed.

This stage bypasses the \gls{rnn} and the two \gls{fc} layers are connected directly to the \gls{cnn}'s output. The procedure thus aims to first train the \gls{cnn} on a simpler task to learn coarse features in terms of a discretised pose representation. The attitude space $\Sothree$ is divided into a spherical grid of discrete azimuth and elevation steps, centred on the target, of fixed radius, i.e.\ a 2-sphere $\Sphere^2$, or viewsphere. Each square on the grid then represents an attitude class $a_i \in \sY_{\Sphere^2}$, ${ i=\{1,\ldots,K_{\Sphere^2}\} }$ with $K_{\Sphere^2}$ possible classes depending on the square size. For the sake of succinctness, the reader is directed to \citet{rondao2021robust} for further details on the viewsphere. The position component is estimated in terms of the relative depth $\lVert \vt \rVert$, thus maximising the joint conditional probability:

\begin{equation}
\vtheta^{\ast\text{(S1)}} = \argmax_{\vtheta^{\text{(S1)}}} p\left(\lVert \vt^{(\dt)} \rVert,\va^{(i,\dt)} \given \tI^{(\dt)};\vtheta^{\text{(S1)}}\right),
\end{equation}

\noindent where $\vtheta^{\text{(S1)}}$ are the \gls{cnn} parameters learned in Stage 1, $\va^{(i)}$ is the one-hot vector encoding of $a_i$, and $\tI$ is the image input. Note that thus far the learning depends only on each individual input at time $\ct=\ct_\dt$, not yet exploiting the temporal correlation in the data.

Sequential images from an \gls{rv} training sequence are preprocessed as follows.
\begin{enumerate*}[label=\arabic*)]
\item First, the attitude space is discretised into the set $\sY_{\Sphere^2}$ with $K_{\Sphere^2}$ classes as mentioned above, discarding any unrepresented class.
\item Define a number $N_{\Sphere^2}$ of desired observations per attitude class.
\item Similarly, $K_{t}$ bins are defined for the relative position $\vt$, selecting the edges according to the minimum and maximum values observed in the dataset, thus creating the set $\sY_{t}$. 
\item For each attitude class $a_c \in \sY_{\Sphere^2}$:
\begin{enumerate*}[label=\arabic{enumi}-\alph*)]
    \item identify the subset  $\sY'_t \subseteq \sY_t$ of $K'_t$ depth bins that contain at least one observation;
    
    \item randomly sample $N_{\Sphere^2}/K'_t$ observations with attitude label $a_c$ equally for each of the $K'_t$ depth bins according to the position ground truth. Oversample if necessary.
\end{enumerate*}
\end{enumerate*}

The resulting Stage 1 dataset will have a total of $N_{\Sphere^2}\cdot K_{\Sphere^2}$ observations with equal representation. For the present application, $N_{\Sphere^2}$ was chosen such that $N_{\Sphere^2}\cdot K_{\Sphere^2} = \num{10000}$. It was found that having balanced attitude classes was paramount to prevent overfitting. To increase data variance in the case of oversampling, an online data augmentation pipeline was implemented, both in terms of visual filtering and small perturbations to the pose.

The loss is formulated as a multi-task learning problem with the attitude component represented by a cross-entropy function and the position component by a regression function, respectively, for each observation $i$:

\begin{align}
    \gL^{(\text{S1})}_{\Sphere^2} &= -\sum\limits_i \sum\limits_{c=1}^{K_{\Sphere^2}} \eva_c^{(i)} \log\left(\hat{\eva}_c^{(i)} \right),\\
    \gL^{(\text{S1})}_t &= \sum\limits_i \dfrac{\lVert \vt^{(i)} - \hat{\vt}^{(i)}\rVert}{\lVert \vt^{(i)}\rVert}, \label{eq:method-losss1_t}
\end{align}

\noindent where $\hat{\va}^{(i)} = \irow{\hat{a}_1^{(i)},\ldots,\hat{a}_{K_{\Sphere^2}}^{(i)}}^\top$ is the predicted attitude class encoding, and $\hat{\vt}^{(i)} \in \sR^3$ is the predicted position. In \gls{vo}, the multi-task loss is typically achieved via linear combination of each component using manually tuned weights; however, as shown by \citet{kendall2018multitask}, this is a sub-optimal approach. Instead, ChiNet models each weight $\{\sigma_{\Sphere^2}, \sigma_t\}$ as learnable task-specific variances of a Boltzmann distribution and a Gaussian distribution, respectively, yielding the combined loss:

\begin{equation}
\gL^{(\text{S1})} = \frac{1}{2} \gL^{(\text{S1})}_{\Sphere^2}  \exp\left(-2 \hat{\sigma}_{\Sphere^2}\right) + \gL^{(\text{S1})}_t \exp\left(-2 \hat{\sigma}_t\right) + \hat{\sigma}_{\Sphere^2} + \hat{\sigma}_t. \label{eq:method-losss1}
\end{equation}

\noindent The reader is directed to \citet{kendall2018multitask} for the details on the derivation\footnote{Despite \eqref{eq:method-losss1_t} not strictly representing the $\normltwo$ component of a Gaussian PDF due to the division by $\lVert \vt^{(i)}\rVert$, the formulation of \eqref{eq:method-losss1} yields good results in practice.} of \eqref{eq:method-losss1}.

\subsection*{Stage 2}

Stage 2 represents ChiNet's nominal training phase of the whole structure, using the normal, non-modified dataset. The full \gls{drcnn} pipeline is trained to maximise the conditional probability of a series of time-sequential poses $\{\mT^{(1)},\ldots,\mT^{(\dt)}\} \in \Sethree$ given a sequence of \gls{rgbt} images, i.e.:

\begin{equation}
\vtheta^{\ast\text{(S2)}} = \argmax_{\vtheta^{\text{(S2)}}} p\left(\mT^{(1)},\ldots,\mT^{(\dt)} \given \tI^{(1)},\ldots,\tI^{(\dt)};\vtheta^{\text{(S2)}}\right), \label{eq:deep-s2maximise}
\end{equation}

\noindent where the \gls{cnn} weights are initialised with the results of Stage 1. Special care must be taken for the representation of the attitude to ensure it remains a member of some group isomorphic to $\Sothree$. A common approach is to admit the unit quaternion representation $\vq$ (e.g.\ \citep{kendall2015posenet,proenca2019deep}) due to the lack of singularities. However, this representation is not continuous due to its antipodal ambiguity (i.e.\ $\vq = -\vq$), which has been shown to introduce learning difficulties into the \gls{dnn} and higher convergence errors.

Instead, ChiNet employs the 6D attitude representation ${\vr \in \sR^6}$ proposed by \citet{zhou2020continuity} which admits a continuous mapping ${\sR^6 \leftarrow \Sothree}$. The transform ${\vr \mapsto \mR}$ entails reshaping $\vr$ into a \num{3x2} matrix followed by Gram-Schmidt orthogonalisation;\footnote{This happens only at inference time and is not needed for training.} the inverse transform thus consists in removing the right-most column of $\mR$. This approach is similar to directly estimating the \num{9} parameters of $\mR$ followed by incorporation of the orthogonalisation procedure inside the network, except with the major advantage of not having to estimate $\num{3}$ superfluous parameters.

The Stage 2 loss is a combined loss based on the $\normltwo$ norm regression of $\vr$ and $\vt$:

\begin{equation}
    \gL^{(\text{S2})}_{r} = \sum\limits_{\dt=1}^T \lVert \hat{\vr}^{(\dt)} - \vr^{(\dt)} \rVert, \   \gL^{(\text{S2})}_{t} = \sum\limits_{\dt=1}^T \lVert \hat{\vt}^{(\dt)} - \vt^{(\dt)} \rVert,
\end{equation}
\begin{dmath}
    \gL^{(\text{S2})} = \gL^{(\text{S2})}_{r}  \exp\left(-2 \hat{\sigma}_{r}\right) + \gL^{(\text{S2})}_t \exp\left(-2 \hat{\sigma}_t\right) + 2\left(\hat{\sigma}_{r} + \hat{\sigma}_t \right), \label{eq:method-losss2}
\end{dmath}

\noindent where \eqref {eq:method-losss2} is derived similarly to \eqref{eq:method-losss1} for two Gaussian distributions, and the temporal component has been highlighted in terms of the training sequence length $T$. Training very long sequences involves high memory requirements, so a truncated \gls{bptt} procedure is adopted instead. This entails unfolding the sequence for a predefined number of time-steps $T$ smaller than the full sequence length $\tilde{T}$, performing one training iteration, and then moving on to the next partition. In order to keep continuity while still allowing the network to learn long sequences, ChiNet follows the approach in \citep{clark2017vinet} whereby the training is carried out with a sliding window over the sequence, where consistency is established by appropriately initialising the \glspl{lstm}'s hidden states with those computed in the previous iteration.

\subsection*{Stage 3}

The final training stage consists in a geometric refinement of the output from Stage 2, following the reprojection of 3D model points using the ground truth and predicted relative pose first proposed by \citet{kendall2017geometric} for camera pose estimation in urban scenarios:

\begin{equation}
\vtheta^{\ast\text{(S3)}} = \argmax_{\vtheta^{\text{(S3)}}} p\left(\mT^{(1)},\ldots,\mT^{(\dt)} \given \tI^{(1)},\ldots,\tI^{(\dt)}, \sP;\vtheta^{\text{(S3)}}\right), \label{eq:method-s3maximise}
\end{equation}

\noindent where $\sP = \{\vp^{(1)},\ldots,\vp^{(N)}\}$ is a manually selected set of $N$ target model points expressed in $\Ft$. The loss is straightforwardly defined as:

\begin{equation}
\gL^{(\text{S3})} = \sum\limits_{\dt=1}^T \sum\limits_{i=1}^N \left\lVert \vz^{(i,\dt)} - \pi\left(\mK,\mT,\vp^{(i)}\right) \right\rVert, \label{eq:method-losss3}
\end{equation}

\noindent where $\sZ_\dt = \{\vz^{(1,\dt)},\ldots,\vz^{(N,\dt)}\}$ is the set of projected keypoints corresponding to $\sP$ at time $\ct=\ct_\dt$, and $\pi$ follows from \eqref{eq:sec-related}. Similarly to Stage 2, the 6D attitude representation is used. \eqref{eq:method-losss3} thus learns the pose implicitly via the minimisation of the reprojection error, which naturally balances the contributions of the position and attitude branches, and does not require defining explicit weights unlike Stages 1 and 2. This is advantageous for datasets in which the position depth has a high variance, since each contribution is weighed differently due to parallax, as reported in \citep{kendall2017geometric}. On the other hand, the loss formulation requires a good initialisation of the parameters $\vtheta^{\text{(S3)}}$ to converge, hence why it is used as a refinement stage.

%% file: s_experiments.tex
\section{Experimental Results}
\label{sec:exp}

\noindent In this section, the performance of the proposed end-to-end \gls{drcnn} pipeline is evaluated on both synthetic and experimental data.

\subsection{Synthetic Dataset}
\label{sec:exp-ssec:synthetic}

\subsubsection{Description}

\begin{figure}[t]
	\centering
	\begin{subfigure}[t]{1\columnwidth}	\centering
	\includegraphics[width=\textwidth]{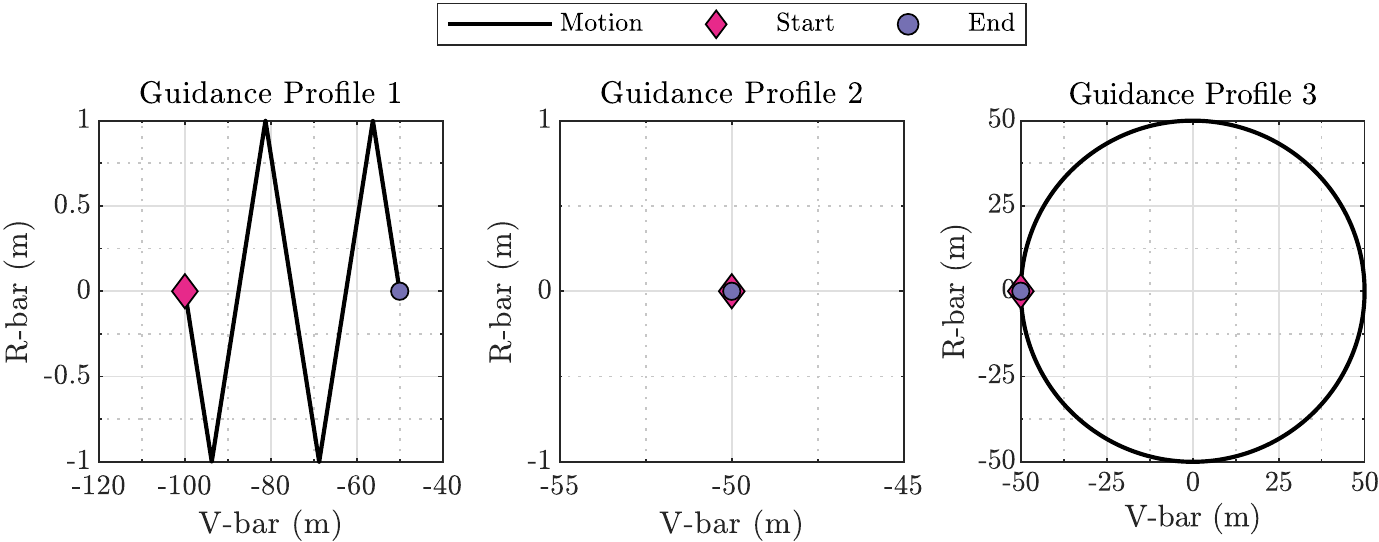}
	\caption{Relative trajectory shapes}
	\label{fig:res-astos-ds-shape}
	\end{subfigure}\\
	\begin{subfigure}[t]{1\columnwidth}	\centering
	\def\svgwidth{\columnwidth}
	\includegraphics[width=\textwidth]{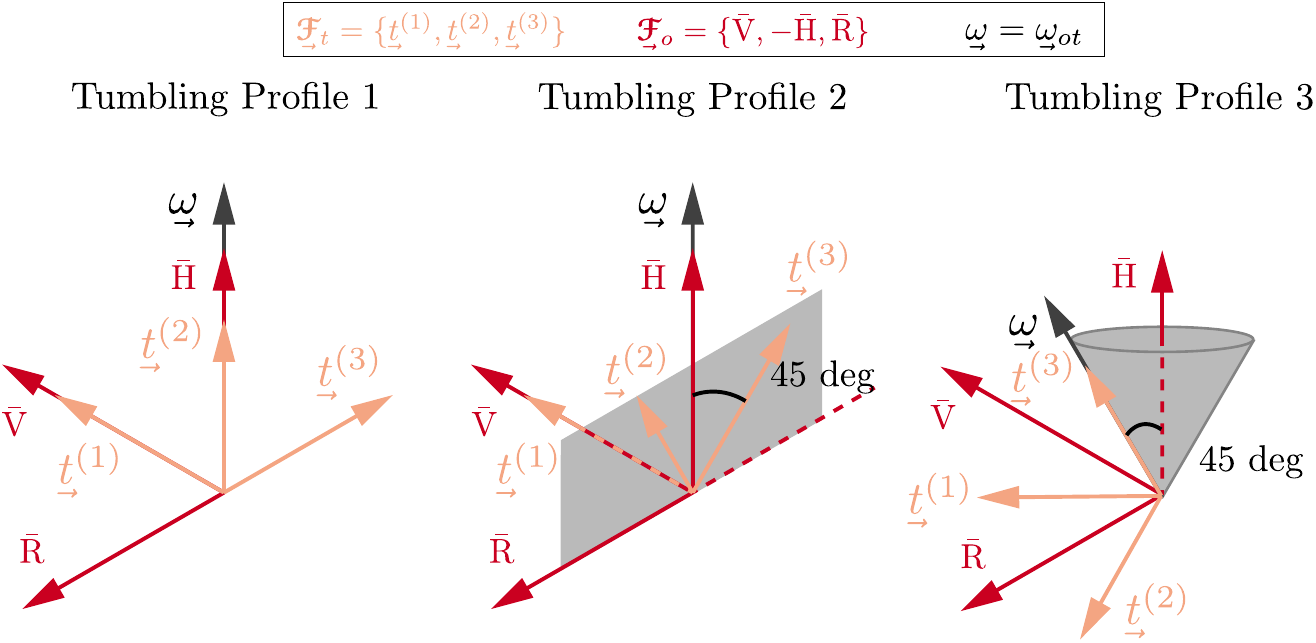}
	\caption{Relative rotational states}
	\label{fig:res-astos-ds-tumbling}
	\end{subfigure}\\
 \caption{Characteristics of the synthetic Astos dataset.}
\label{fig:res-astos-ds}
\end{figure}

The framework is initially validated on the Astos dataset, consisting of \num{14} different rendezvous trajectories with the failed satellite Envisat, featuring three distinct \glspl{gp}, three tumbling modes, and two approach vectors. The images are synthetically generated using the Astos Camera Simulator\footnote{\url{https://www.astos.de/products/camsim}.} with emulated visible and thermal cameras at a frequency of \SI{10}{\hertz}. The visible and \gls{lwir} images are aligned and resized to a resolution of \SI{640x512}{\pixel} for both training and testing. The reader is directed to \citet{rondao2020benchmarking} for details on the chosen Envisat orbital parameters and image generation. \Figref{fig:res-astos-ds-shape} illustrates the three considered \glspl{gp} of the chaser expressed in the target's \gls{lvlh} frame $\Fo$ \citep{fehse2003automated}. \Figref{fig:res-astos-ds-tumbling} depicts the considered rotational states for Envisat; a note is made relative to \gls{tp} 3, in which the spin axis is configured at a \SI{45}{\degree} angle with H-bar but is simultaneously fixed in the inertial frame. Since Envisat’s orbit is approximately circular, this results in the spin axis demonstrating an axial precession with period equal to the orbital period, or \SI{3.59}{\degree\per\minute}. Apart from the guidance and tumbling profiles, additional variation is added via the approach vector (see \tabref{tab:res-astos-seq}), where the V-bar case features a black, deep-space background, and the R-bar case contains Earth in the \gls{fov}.

\subsubsection{Training and Testing}

\begin{table}[t]
\caption{Sequence key for the Astos dataset.}
\label{tab:res-astos-seq}
\begin{tabularx}{\columnwidth}{@{}l *{4}{C} c@{}}
\toprule
\textbf{Sequence}        & \textbf{GP}    & \textbf{TP}    & \textbf{Approach Vector}   & \textbf{Selection}     & \textbf{Length (\si{\second})}\\ 
\midrule
\texttt{00}     & 1     & 1     & V-bar             & Train         &   125\\
\texttt{01}     & 1     & 1     & R-bar             & Test          &   125\\
\texttt{02}     & 1     & 2     & V-bar             & Test          &   125\\
\texttt{03}     & 1     & 2     & R-bar             & Train         &   125\\
\texttt{04}     & 1     & 3     & V-bar             & Train         &   125\\
\texttt{05}     & 1     & 3     & R-bar             & Test          &   125\\
\texttt{06}     & 2     & 1     & V-bar             & Test          &   309\\
\texttt{07}     & 2     & 1     & R-bar             & Train         &   309\\
\texttt{08}     & 2     & 2     & V-bar             & Train         &   216\\
\texttt{09}     & 2     & 2     & R-bar             & Test          &   216\\
\texttt{10}     & 2     & 3     & V-bar             & Test          &   216\\
\texttt{11}     & 2     & 3     & R-bar             & Train         &   216\\
\texttt{12}     & 3     & 1     & N/A               & Train         &   200\\
\texttt{13}     & 3     & 2     & N/A               & Test          &   200\\
\bottomrule
\end{tabularx}
\end{table}

A train-test split is performed on the Astos dataset according to \Tabref{tab:res-astos-seq}, where one half of the sequences are used for training and the other half for testing. The split was performed so that the network is trained at least once on each \gls{gp} and \gls{tp}, but the tests include different combinations thereof.

The sequences are further partitioned for training according to randomly sampled lengths of $\{64,128,256,512\}$ \si{\second}. \citepos{clark2017vinet} method is used to train the \gls{rnn} module whereby each sequence is fed to the network according to a sliding window. In the present experiments, a window length of \num{8} frames with a stride of \num{4} was utilised.

Image augmentation is performed online (i.e.\ during training) on the data in terms of image processing (e.g.\ random brightness and contrast, Gaussian blur and noise, random pixel dropout, etc.) and camera perturbations by manipulating the image according to a homography computed through a pure rotation.

Stages \num{1} and \num{2} are trained for \num{100} epochs with a cyclical learning rate decay of \num{5} cycles, whereas Stage \num{3} is trained for \num{66} epochs with early stopping and a step learning rate decay every \num{9} epochs. Stage \num{1} samples the dataset for a total of \num{10000} images. The \gls{cnn} and \gls{rnn} modules are trained separately, but sequentially. The Adam optimiser \citep{kingma2014adam} is used. The final pipeline uses a dropout probability of \num{0.2}, and hidden and cell states zoneout factors of \num{0.15} for both. 

The \gls{drcnn} is implemented from the ground up on MATLAB version R2019b. The pipeline is trained one NVIDIA\textregistered\ Turing\textregistered\ V100 Tensor Core \gls{gpu} with a minibatch size of \num{128}.

\subsubsection{Evaluation}

The test results are presented in terms of the position and attitude error metrics, respectively:

\begin{align}
\delta\tilde{t} &\coloneqq \lVert \hat{\vt} - \vt \rVert,\\
\delta\tilde{q} &\coloneqq 2\arccos\left(\hat{\vq}^{-1} \otimes \vq\right)_4,
\end{align}

\noindent where $\hat{\bullet}$ denotes the estimated quantity, $\otimes$ denotes quaternion multiplication, and the subscript ``\num{4}'' refers to the scalar element of the quaternion. is Additionally, the position error is also assessed in terms of the relative range:

\begin{equation}
\delta\tilde{t}_\urr \coloneqq \dfrac{\delta\tilde{t}}{\lVert \vt \rVert}.
\end{equation}

\noindent For succinctness, the \texttt{ASTOS/06} sequence is used as a representative case study, where the errors are plotted as a function of time, whereas the results for the remaining sequences are summarised for the complete pipeline in terms of their mean and median statistics. 

\subsection*{Evaluation of Multistage Optimisation}

\begin{figure}[t]
  	\centering
  		\includegraphics[width=\columnwidth]{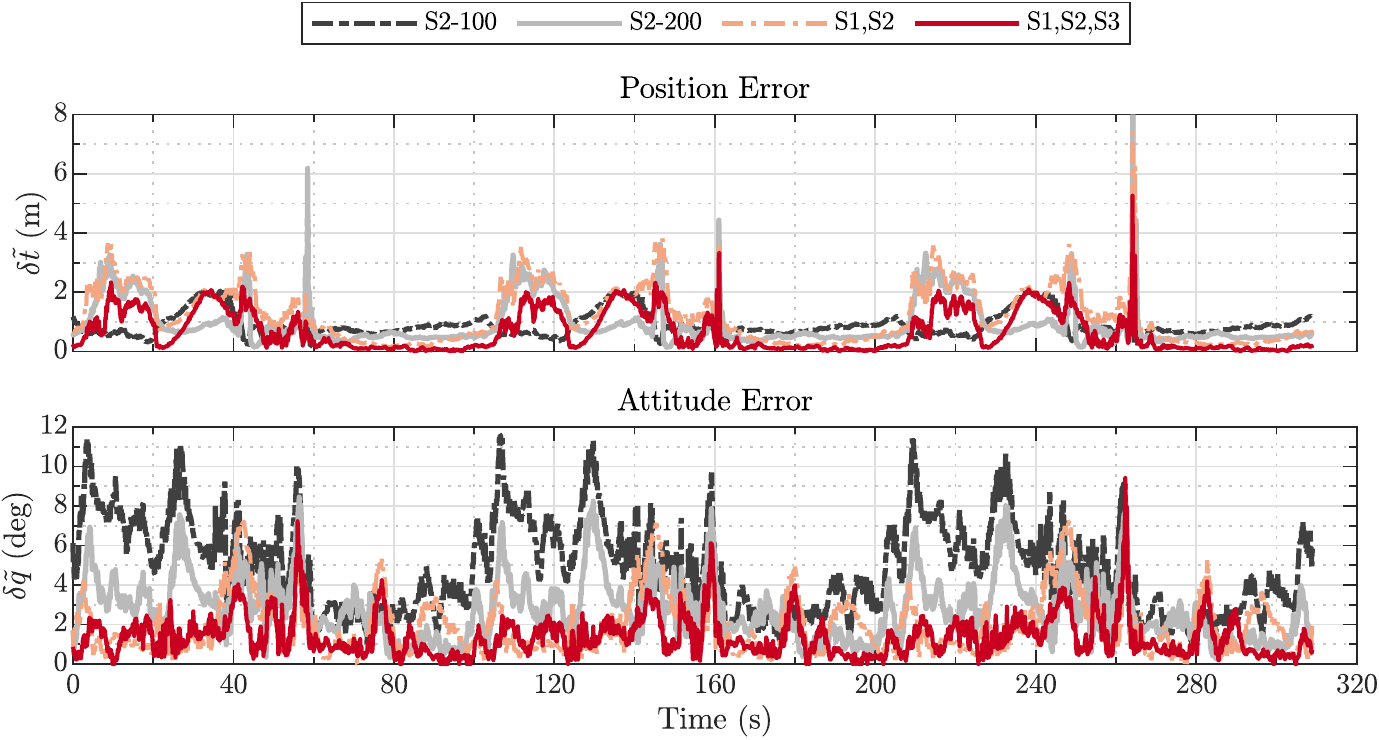}
  		\caption{Comparison of estimated position and attitude errors over time in terms of training stages used for \texttt{ASTOS/06}. All models are trained on a \glsxtrshort{cnn} taking \glsxtrshort{rgb} inputs. \figcustom{S2-100} Stage \num{2} trained for \num{100} epochs. \figcustom{S2-200} Stage \num{2} trained for \num{200} epochs. \figcustom{S1,S2} Stage \num{1} and Stage \num{2}. \figcustom{S1,S2,S3} Stage \num{1}, Stage \num{2}, and Stage \num{3}.}
        \label{fig:deep-res-test01}
\end{figure}

To assess each contribution in the proposed multistage optimisation scheme, the \gls{cnn} module on its own is first considered, and trained according to four different schemes:
\begin{enumerate*}[label=\arabic*)]
\item Stage \num{2} only for \num{100} epochs [S2-100];
\item Stage \num{2} only for \num{200} epochs [S2-200];
\item Stages \num{1} and \num{2} [S1,S2]; and
\item Stages \num{1}, \num{2}, and \num{3} [S1,S2,S3].
\end{enumerate*}
The \gls{rnn} is not considered for this test.

{
\sisetup{
  range-phrase = \ {;}\ ,
  range-units  = brackets,
  open-bracket = [,
}
\Figref{fig:deep-res-test01} depicts the results of the benchmark on the baseline. From the overall shape of the plot lines, the periodicity of the tumbling motion can be clearly discerned. An initial period approximately covering the interval $\tau \in \SIrange[close-bracket= {[}]{0}{60}{\second}$ is first noted, during which the target performs slightly over half a revolution and the errors are overall higher, culminating in a local peak at which the solar array reflects Earth's rim. It is then followed by a second period covering $\tau = \SIrange[close-bracket= {[}]{60}{103}{\second}$ where the main body (also known as ``bus'') comes back into view and both shadows and reflections are minimised, hence driving down the errors. This pattern is repeated twice more throughout the plot as the target performs a total of three revolutions.
}

Regarding the position error, the S1,S2 strategy is essentially on par with S2-100 and S2-200 for the first period, and performs better than both on the second period. Notably, the benefit of the dual-stage training can be observed specifically at times $\tau = \{\num{60},\num{160},\num{260}\}\,\si{\second}$, where a mitigation of the error spikes is seen. Training on the three stages (S1,S2,S3) reduces these peaks even further.

The gains of adopting the proposed method become clearer looking at the attitude error plot. S2-100 exhibits the higher error throughout, followed by S2-200. The dual-stage S1,S2 approach further reduces the error, except for peaks at \{\num{45},\num{147},\num{250}\} \si{\second}, where it is comparable to the previous mode; this corresponds to the segments where the target nearly completes half a revolution and the solar array begins to cover the main bus. The triple-stage approach can be seen to provide the steadiest performance. It is also noted that the highest error peaks for the attitude correspond to those identified for the position, which S1,S2,S3 mitigates, but does not completely eliminate.

\subsection*{Evaluation of Recurrent Module}

\begin{figure}[t]
  	\centering
  		\includegraphics[width=\columnwidth]{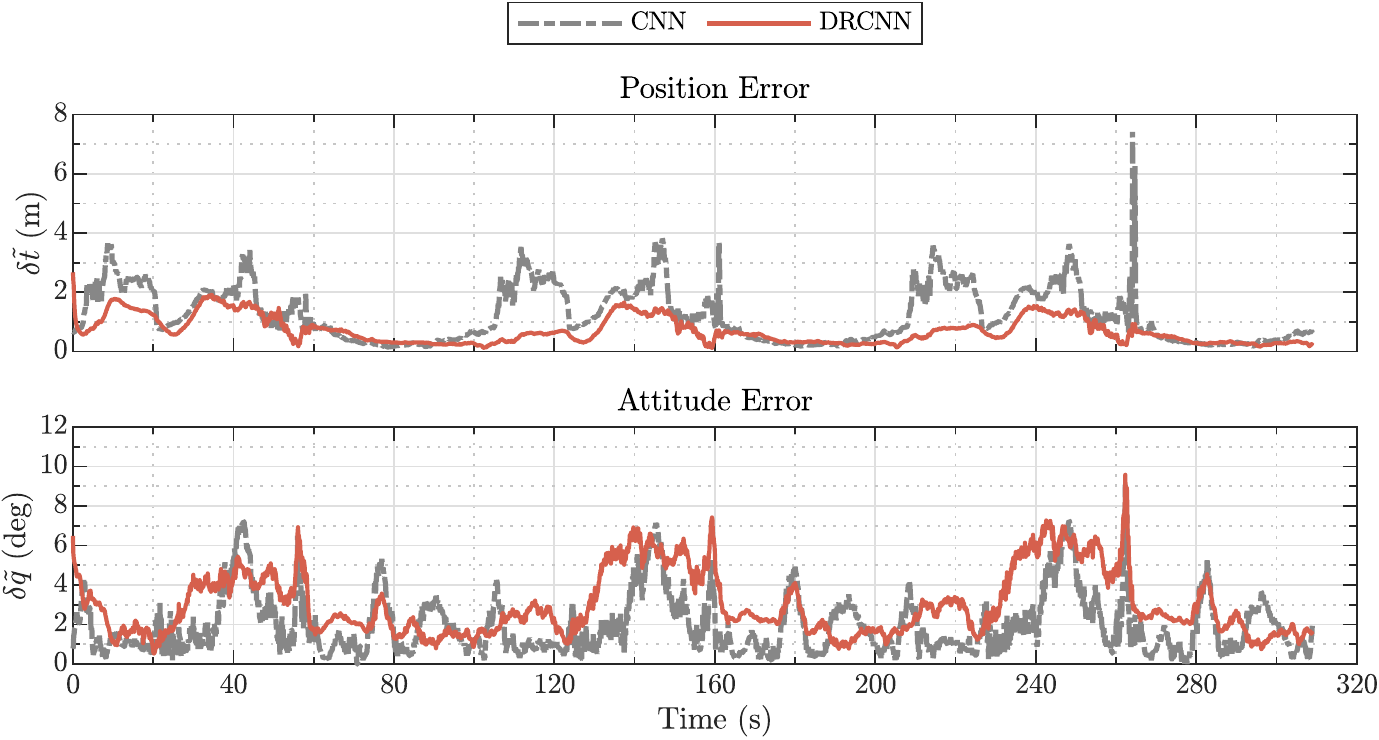}
  		\caption{Comparison of estimated position and attitude errors over time in terms of recurrence for \texttt{ASTOS/06}, benchmarking the plain \glsxtrshort{cnn} against the complete \glsxtrshort{drcnn}. All models are trained on Stages \num{1} and \num{2} and \gls{rgb} inputs.}
        \label{fig:deep-res-test02}
\end{figure}

In this section, the performance of the \gls{cnn} is compared to the complete \gls{drcnn}; \Figref{fig:deep-res-test02} plots the estimation results over time, where the training regime consisted of S1,S2, and \gls{rgb} inputs are considered. The \gls{drcnn} is successful in overwhelmingly mitigating the localised position error peaks, which correspond to points in the trajectory where the solar array reflections are most intense or it occludes the main bus, as mentioned in the previous section. This is due to the \gls{lstm} states taking into account the preceding images, thus preventing sudden jumps in the solution. The mean position error is reduced approximately by half, bringing the mean range-normalised error to approximately \SI{1.40}{\percent}.

The mean values for the attitude errors, however, are slightly worse for the \gls{rnn}-based architecture. Overall, an increase of \SIrange{0.5}{1}{\degree} in the mean error and \SI{1}{\degree} in the median error is observed. It can be argued that this is an acceptable loss in performance given the benefit seen for the position estimation. However, the pipeline could instead be modified to output an attitude estimate from the \gls{cnn} alone while processing the position with the \gls{rnn}. This is left as future work.

\subsection*{Evaluation of Multimodal Inputs}

\begin{figure}[t]
  	\centering
  		\includegraphics[width=\columnwidth]{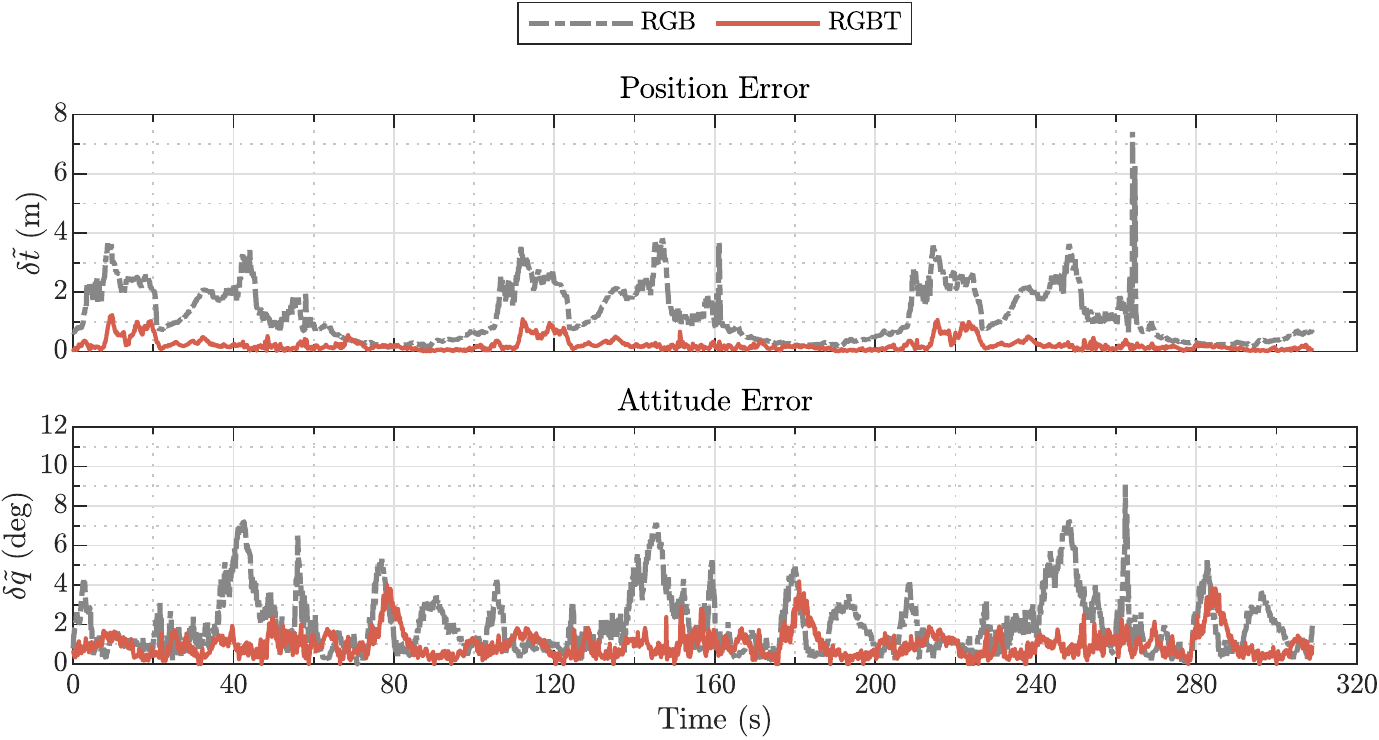}
  		\caption{Comparison of estimated position and attitude errors over time in terms of imaging modality for \texttt{ASTOS/06}, benchmarking \glsxtrshort{rgb} inputs against the multimodal \glsxtrshort{rgbt}. All models are trained on a \glsxtrshort{cnn} and Stages \num{1} and \num{2}.}
        \label{fig:deep-res-test03}
\end{figure}

In this section, the influence of augmenting the \gls{rgb} input produced by regular camera with an image in the \gls{lwir}, thus creating a four channel multimodal \gls{rgbt} input, is assessed. Two models are trained for comparison, one with inputs exclusively on the visible modality, and another with multimodal inputs. Both models are trained on Stages \num{1} and \num{2}. Again, the \gls{rnn} is not considered for this test so as to separate the effect of each contribution. The results are depicted in \Figref{fig:deep-res-test03}.

The contribution of the multimodality can be seen immediately from the figure, where the plots of both position and attitude errors in time exhibit more stability for \gls{rgbt} inputs compared to \gls{rgb} inputs. Notably, not only are the reflection-induced peaks mitigated, but the errors corresponding to the approximate first half of the tumbling period are as well. Overall, the mean position error is reduced in almost \SI{80}{\percent} by using multimodal inputs, granting a mean range-normalised position error below \SI{0.5}{\percent}, compared to \SI{2.5}{\percent} for visible only. The mean attitude error is halved, becoming slightly lower than \SI{1}{\degree}.

\subsection*{Summary of Performance}

\begin{figure}[t]
  	\centering
  		\includegraphics[width=\columnwidth]{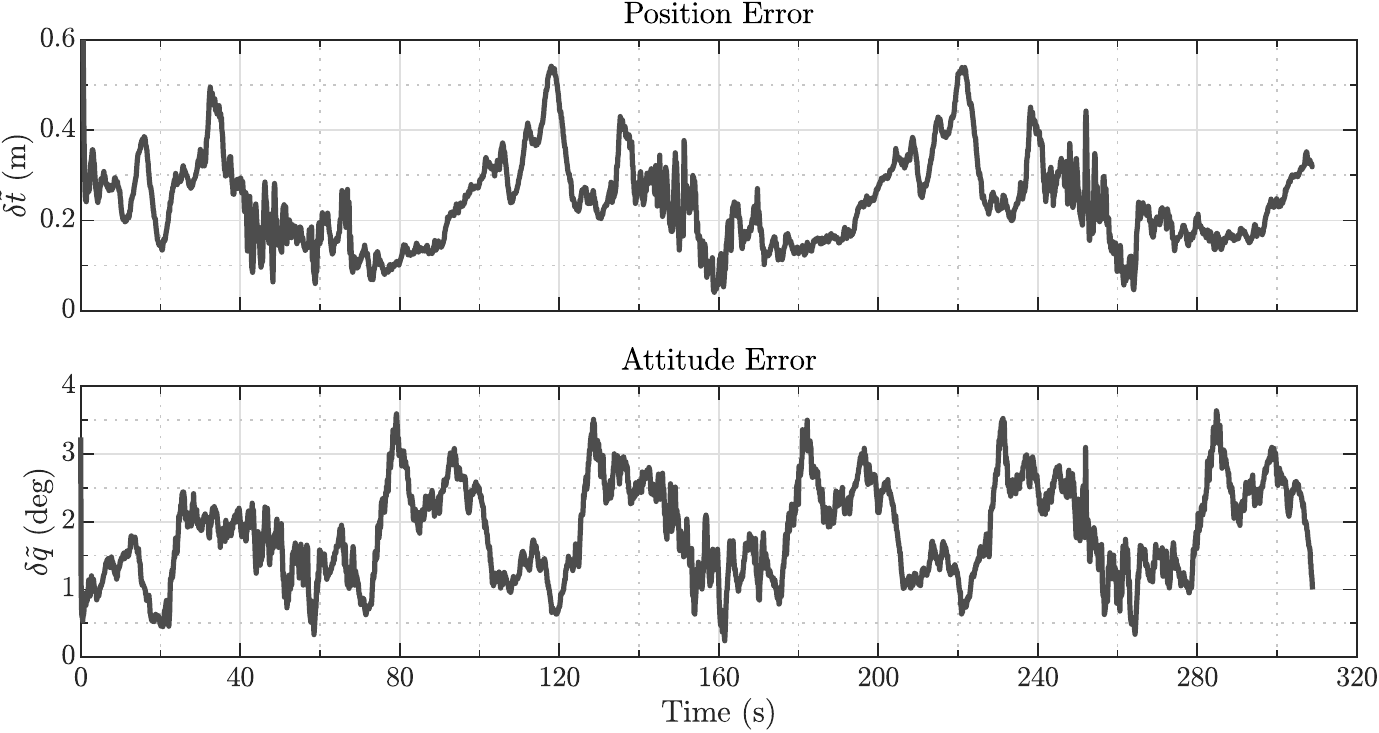}
  		\caption{Estimated position and attitude errors over time on the \texttt{ASTOS/06} rendezvous sequences for the complete multimodal \glsxtrshort{drcnn}.}
        \label{fig:deep-res-test00}
\end{figure}

\Tabref{tab:deep-res-test00} compiles the error statistics for the performance of the complete multimodal \gls{drcnn} framework on the entire Astos test dataset. For completeness, the performance on the nominal sample sequence is also benchmarked in \Figref{fig:deep-res-test00}, and illustrated qualitatively in \Figref{fig:intro-res-frames-test00-full-06} (\figref{fig:intro-res-frames-test00-full-13} showcases the performance on \texttt{ASTOS/13}).

\begin{table}[t]
\sisetup{detect-weight=true,detect-inline-weight=math}
\centering
 \caption{Summary of position and attitude error statistics on all Astos dataset rendezvous test sequences for the complete \glsxtrshort{drcnn} pipeline, trained on Stages 1, 2, and 3. All tests use multimodal \glsxtrshort{rgbt} inputs.}
 \label{tab:deep-res-test00}
 \begin{adjustbox}{width={\columnwidth},totalheight={\textheight},keepaspectratio}
 \begin{tabular}{@{}l S S S S S S} 
	\toprule
	\multirow{2}{*}{\textbf{Sequence}} & \multicolumn{2}{c}{$\delta\tilde{t}$ \textbf{(\si{\metre})}} & \multicolumn{2}{c}{$\delta\tilde{t}_\urr$ \textbf{(-)}} & \multicolumn{2}{c}{$\delta\tilde{q}$ \textbf{(\si{\degree})}}\\
	&	{\textbf{Mean}}	& {\textbf{Median}}	&	 {\textbf{Mean}}	&	{\textbf{Median}}	&	{\textbf{Mean}}	&	{\textbf{Median}}\\
	\midrule
	\texttt{01} & 3.45 & 3.51 & 0.0468 & 0.0473 & 7.49 & 4.80\\
	\texttt{02} & 4.05 & 4.23 & 0.0547 & 0.0581 & 8.67 & 4.53\\
	\texttt{05} & 3.09 & 3.12 & 0.0437 & 0.0437 & 14.12 & 8.63\\
    \texttt{06} & 0.24 & 0.23 & 0.0049 & 0.0046 & 1.85 & 1.80\\
	\texttt{09} & 0.33 & 0.24 & 0.0065 & 0.0048 & 2.09 & 1.26\\
	\texttt{10} & 0.67 & 0.63 & 0.0134 & 0.0126 & 10.61 & 9.02\\
	\texttt{13}& 0.29 & 0.21 & 0.0058 & 0.0041 & 3.52 & 2.77\\
	\bottomrule
 \end{tabular}
 \end{adjustbox} 
\end{table}

The performance of ChiNet can be directly compared to the classic \acrconnect{ml}{-based} algorithm developed by the authors in \citep{rondao2021robust} (herein referred to as ``classical'') through the \texttt{ASTOS/06} and  \texttt{ASTOS/02} sequences, since these have also been considered for that analysis. Starting with the first one, it can be seen that ChiNet provides an estimate of the position with an error bound at \SI{0.6}{\metre}, scoring on average a mean $\delta\tilde{t}_\urr=\SI{0.49}{\percent}$. The classical solution, on the other hand, reached maximum values of \SI{2.5}{\metre}. For this trajectory, ChiNet presents an improvement of around \num{2.2} percentage points in terms of mean range-normalised position error. The classical solution performs better in terms of mean attitude error (\SI{0.78}{\degree}). Still, ChiNet produces a solution not exceeding \SI{2}{\degree} in error.

Considering the remaining sequences within \gls{gp}2 (fixed relative range), it can be seen that the quality of the solution degrades as more challenging rotation modes are considered. The estimation of the attitude appears to be more affected by this factor. For mode \gls{tp}2 (two-axis rotation), the pose errors are comparable to \gls{tp}1, even despite the benchmark of the former being performed on an R-bar approach vector (i.e.\ with Earth in the \gls{fov}). Mode \gls{tp}3 (precession) experiences by far the largest degradation, with the mean attitude error exceeding \SI{10.5}{\degree}. On sequences featuring this rotation mode, the edge of the solar array leaves the \gls{fov} for a considerable amount of time, which could explain the higher error.

Overall, \gls{gp}1 trajectories (forced translation) exhibit reduced performance when compared to \gls{gp}2. This was expected since the network sees far more examples of the relative pose at a distance of \SI{50}{\metre} than at larger distances. Nevertheless, for this profile ChiNet produces estimates of the position with mean $\delta\tilde{t}_\urr$ not exceeding \SI{5.5}{\percent}. The mean attitude error is less affected by the change in guidance profile, being \SIrange[range-units=single]{1.5}{4}{\texttimes} higher with respect to \gls{gp}2. Taking \texttt{ASTOS/02} as an example, the mean $\delta\tilde{t}_\urr=\SI{5.47}{\percent}$ is approximately \num{2.7} percent points higher than the output of the classical algorithm. The mean attitude error is also higher (\SI{3.4}{\texttimes}).

\subsection{Experimental Dataset}

\subsubsection{Description}

Lastly, the performance of the complete ChiNet pipeline is assessed on real data acquired from the \gls{asmil} at City, University of London (herein referred to as ``City dataset''). This test provides insight on how well the deep learning framework can adapt to data captured by actual sensors, and to the sources of error a laboratory setup brings (e.g. camera calibration; ground truth measurement; camera misalignments; camera synchronisation; sensor noise). It also evaluates how the network fares against previously unseen motion when trained on reduced amounts of data. 

\begin{figure}
  	\centering
	  	\begin{subfigure}[t]{0.24\columnwidth}\centering
  		\includegraphics[trim=100 100 100 100,clip,width=\textwidth]{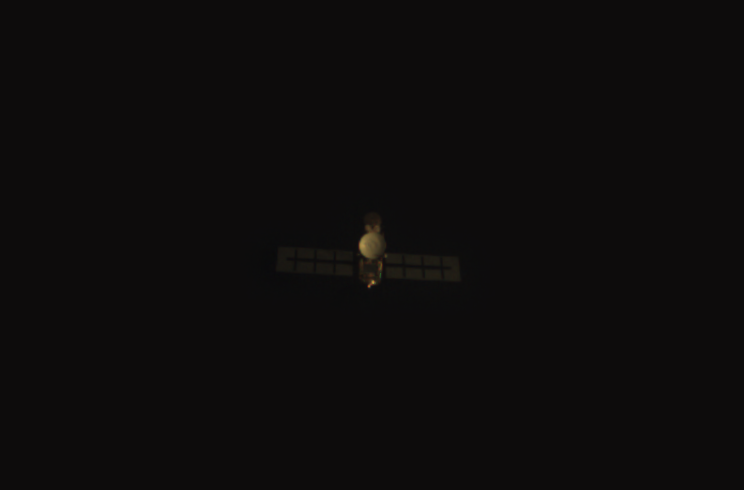}
	\end{subfigure}
	  	\begin{subfigure}[t]{0.24\columnwidth}\centering
  		\includegraphics[trim=100 100 100 100,clip,width=\textwidth]{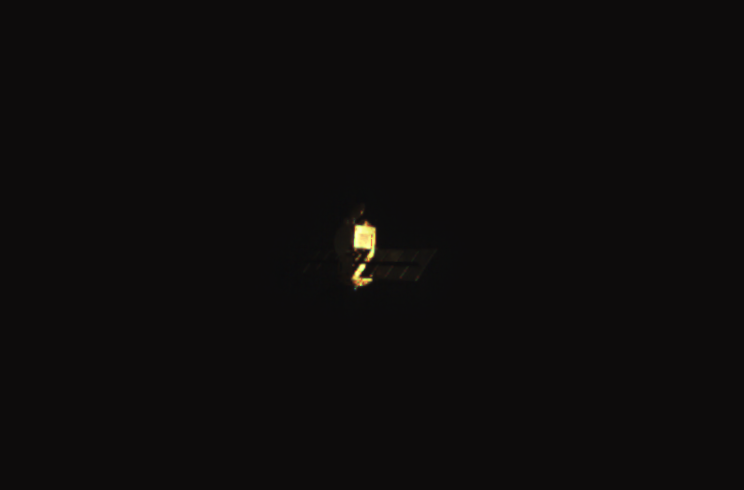}
  		\end{subfigure}
  		 \begin{subfigure}[t]{0.24\columnwidth}\centering
  		 \includegraphics[trim=100 100 100 100,clip,width=\textwidth]{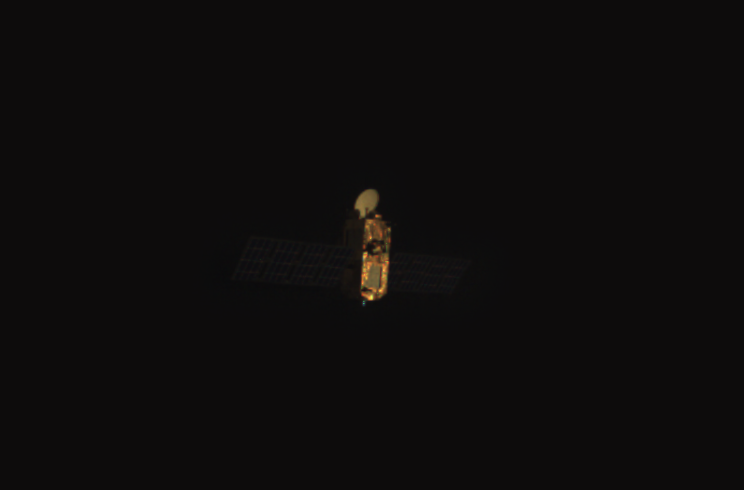}
  		\end{subfigure}
  		\begin{subfigure}[t]{0.24\columnwidth}\centering
  		\includegraphics[trim=100 100 100 100,clip,width=\textwidth]{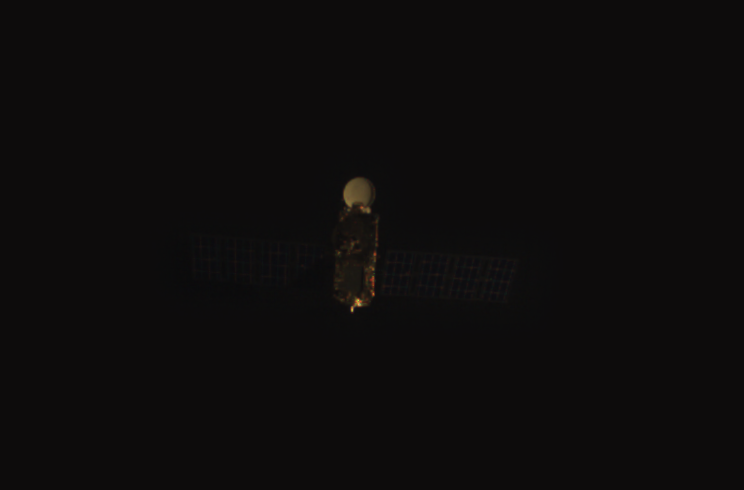}	
	\end{subfigure}\\ \vspace{1em}
	  	\begin{subfigure}[t]{0.24\columnwidth}\centering
  		\includegraphics[trim=100 100 100 100,clip,width=\textwidth]{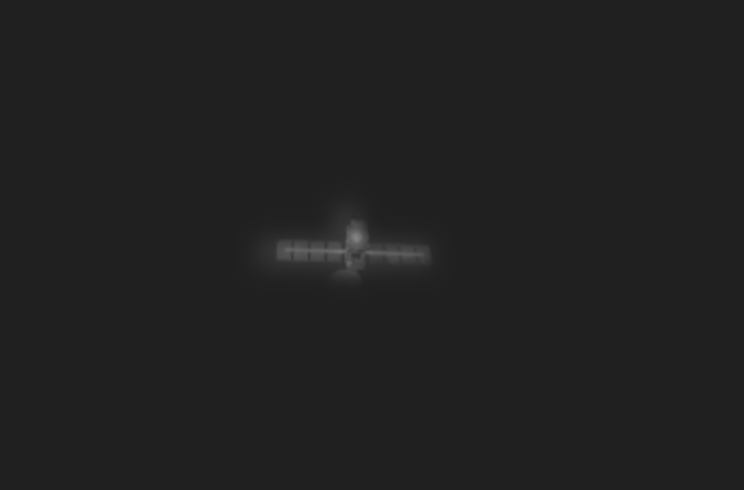}
	\end{subfigure}
	  	\begin{subfigure}[t]{0.24\columnwidth}\centering
  		\includegraphics[trim=100 100 100 100,clip,width=\textwidth]{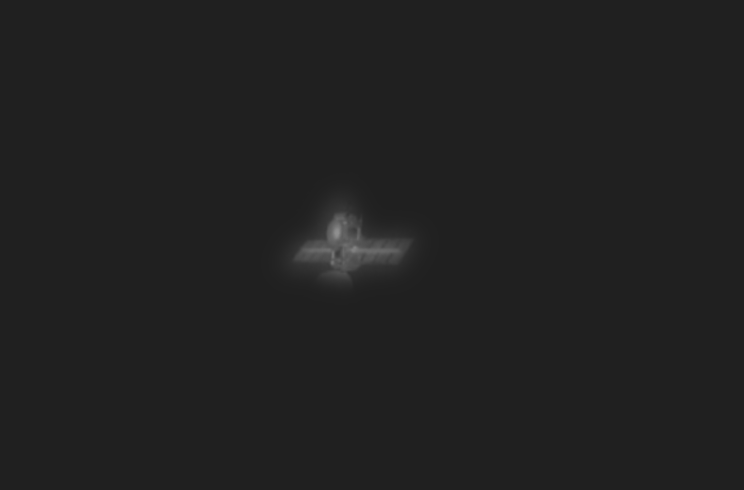}
  		\end{subfigure}
  		 \begin{subfigure}[t]{0.24\columnwidth}\centering
  		 \includegraphics[trim=100 100 100 100,clip,width=\textwidth]{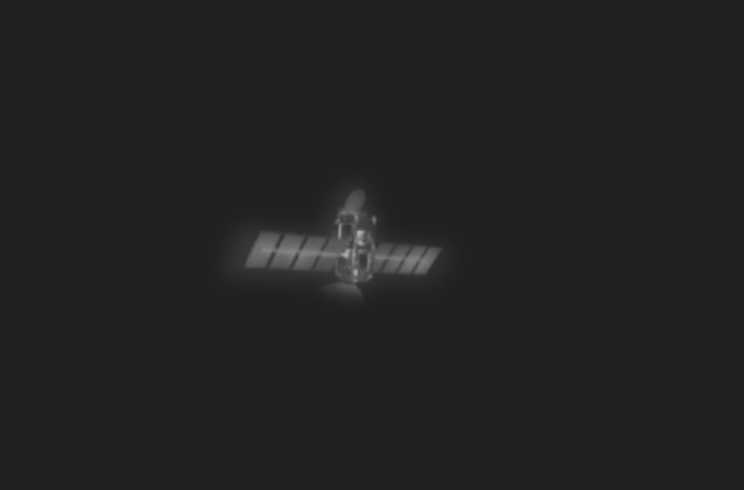}
  		\end{subfigure}
  		\begin{subfigure}[t]{0.24\columnwidth}\centering
  		\includegraphics[trim=100 100 100 100,clip,width=\textwidth]{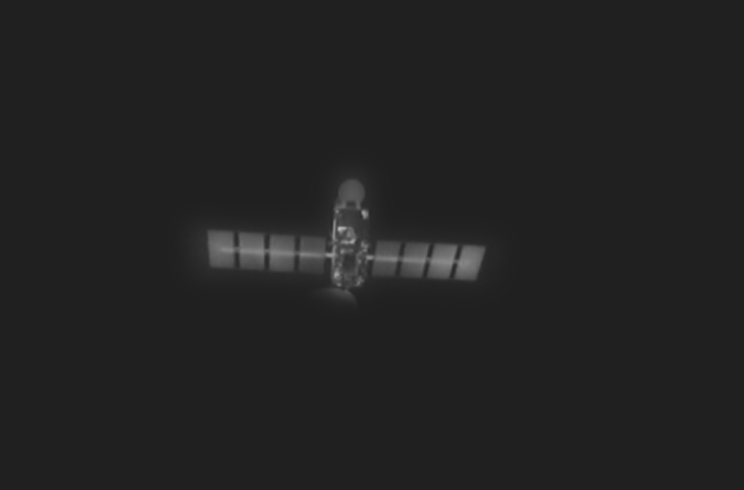}	
	\end{subfigure}			
  \caption{Sample images from the City dataset (cropped for visualisation purposes). \figcustom{Top row} Visible modality. \figcustom{Bottom row} \glsxtrshort{lwir} modality. }
  \label{fig:deep-res-showcase-city}
\end{figure}

The City dataset consists of a multimodal collection of four rendezvous sequences with a \num{1}:\num{4} scale mock-up of the Jason-1 satellite. The mock-up rotates along its vertical axis at a constant rate of \SI{6}{\degree\per\second}. Despite having a different form factor, Jason-1 is similar to Envisat in terms of components (i.e.\ main bus coated in \gls{mli}, thermal radiators, solar array, radiometric instruments). In total, four trajectory types are considered. \Tabref{tab:res-city-seq} summarises the characteristics of each sequence, and \Figref{fig:deep-res-showcase-city} shows some sample images from the dataset in each modality.

\begin{table}[t]
\caption{Sequence key for the City dataset.}
\label{tab:res-city-seq}
\begin{tabularx}{\columnwidth}{@{}l *{4}{C} c@{}}
\toprule
Sequence        & GP    & Initial dist. (\si{\metre})    & Final dist. (\si{\metre})   & Rotation (\si{\revolution}) & Length (\si{\second})\\ 
\midrule
\texttt{00}     & Fixed      & 3.8     & 3.8             & 2         &   120\\
\texttt{01}     & Fixed      & 1.1     & 1.1             & 2          &   120\\
\texttt{02}     & Translation     & 3.8     & 1.1        & 0.5          &   30\\
\texttt{03}     & Translation     & 3.8    & 2           & 0.5         &   30\\
\bottomrule
\end{tabularx}
\end{table}

\begin{figure}[t]
  	\centering
  	\includegraphics[width=0.6\columnwidth]{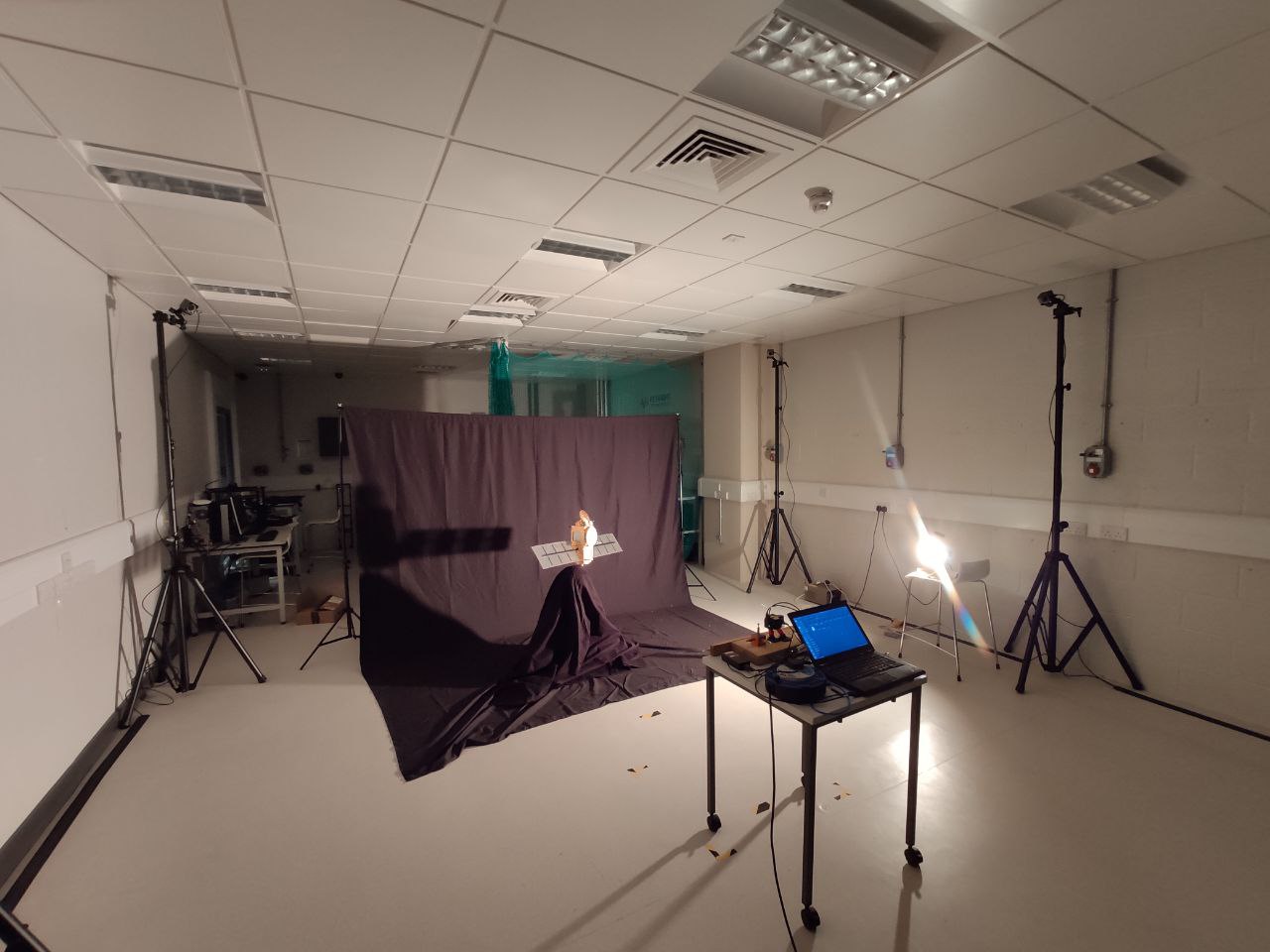}
    \caption{Validation setup of the \glsxtrshort{asmil} at City, University of London.}
  \label{fig:intro-setup-asl}
\end{figure}

\noindent Trajectories are acquired for simulation of both sunlight and eclipse conditions. On the visible spectrum, this is controlled respectively by aiming a floodlight directly at the target, or by aiming it at a nearby wall, creating a dimly lit environment. On the \gls{lwir} spectrum, the model's temperature is controlled by internal resistor heaters in the main bus and by an external heater. The thermal signature of the model is made to coarsely match that of Envisat in both illumination conditions. Images are acquired at a resolution of \SI{744x490}{\pixel} and frequency of \SI{10}{\hertz} (software synchronised). The visible and thermal cameras are aligned and set up in a stereo configuration with a very short baseline to minimise disparity. The ground truth is recorded with an six-camera Optitrack\footnote{\url{https://optitrack.com}.} motion caption system. Using the ground truth and the \gls{cad} model of the target, the background is digitally masked out to simulate a deep space background. \Figref{fig:deep-res-showcase-city} depicts some sample frames of the dataset, whereas   \Figref{fig:intro-setup-asl} showcases the experiment setup at \gls{asmil}.

\subsubsection{Training and Testing}

The methodology follows analogously from \Secref{sec:exp-ssec:synthetic}. The pipeline is trained on \texttt{CITY/00}, \texttt{CITY/01}, and \texttt{CITY/02}, and is evaluated on \texttt{CITY/03}.

\bigskip

\begin{figure}
  	\centering
  		\includegraphics[width=\columnwidth]{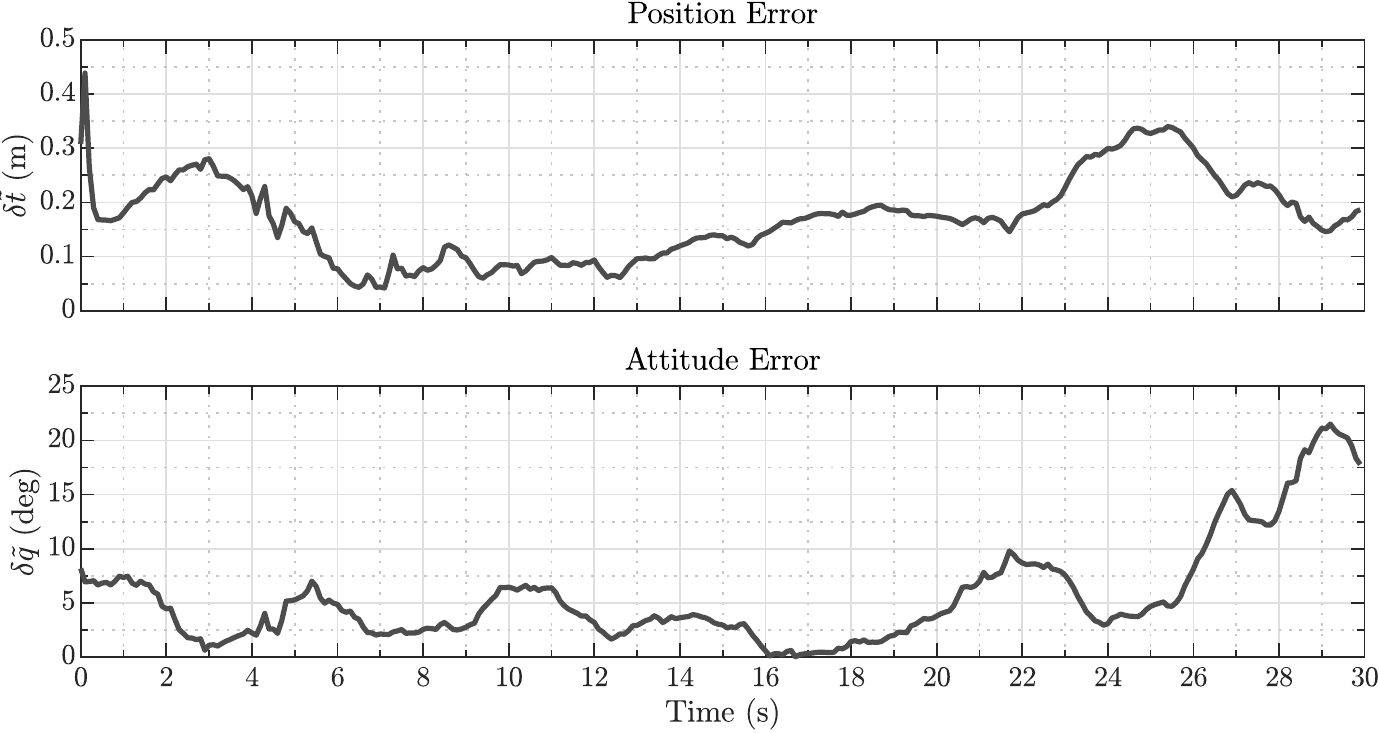}
  \caption{Estimated position and attitude errors over time on the \texttt{CITY/03} laboratory test rendezvous sequences. The model is trained on the full \glsxtrshort{drcnn} pipeline with multimodal \glsxtrshort{rgbt} inputs and on Stages \num{1}, \num{2}, and \num{3}.}
  \label{fig:res-test04}
\end{figure}

\begin{figure}[t]
  	\centering
  	\begin{subfigure}[t]{\columnwidth}\centering
  		\includegraphics[trim=125 100 125 100,clip,width=0.49\textwidth]{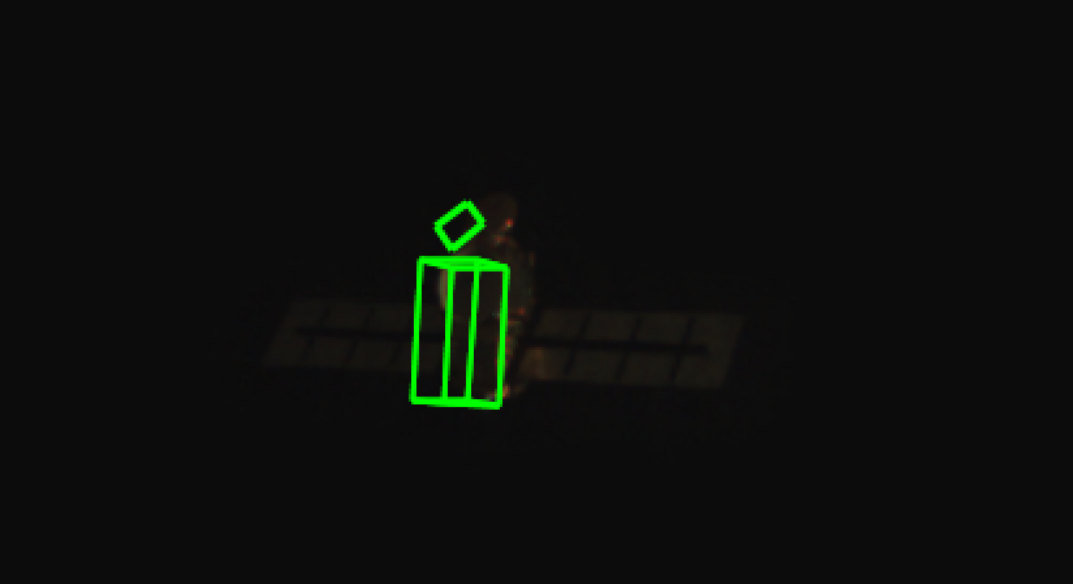}
  		\includegraphics[trim=125 100 125 100,clip,width=0.49\textwidth]{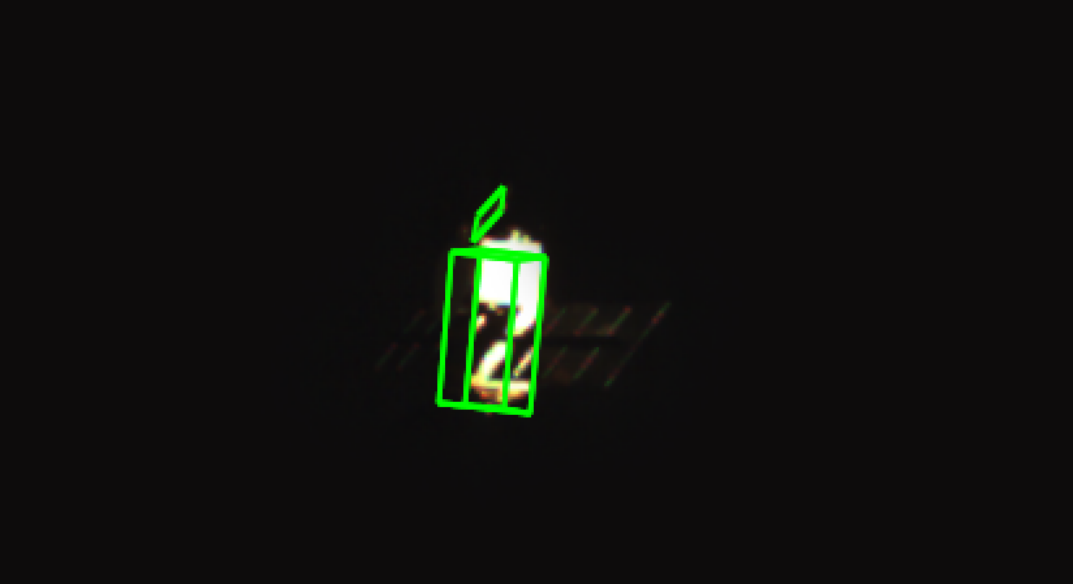}\\ \vspace{1em}
  		\includegraphics[trim=125 100 125 100,clip,width=0.49\textwidth]{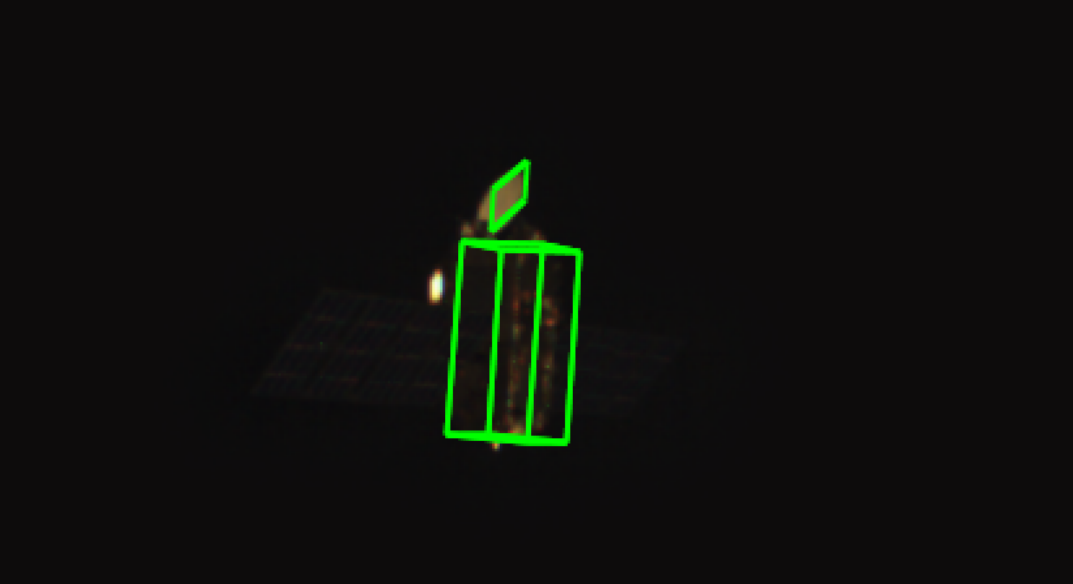}
  		\includegraphics[trim=125 100 125 100,clip,width=0.49\textwidth]{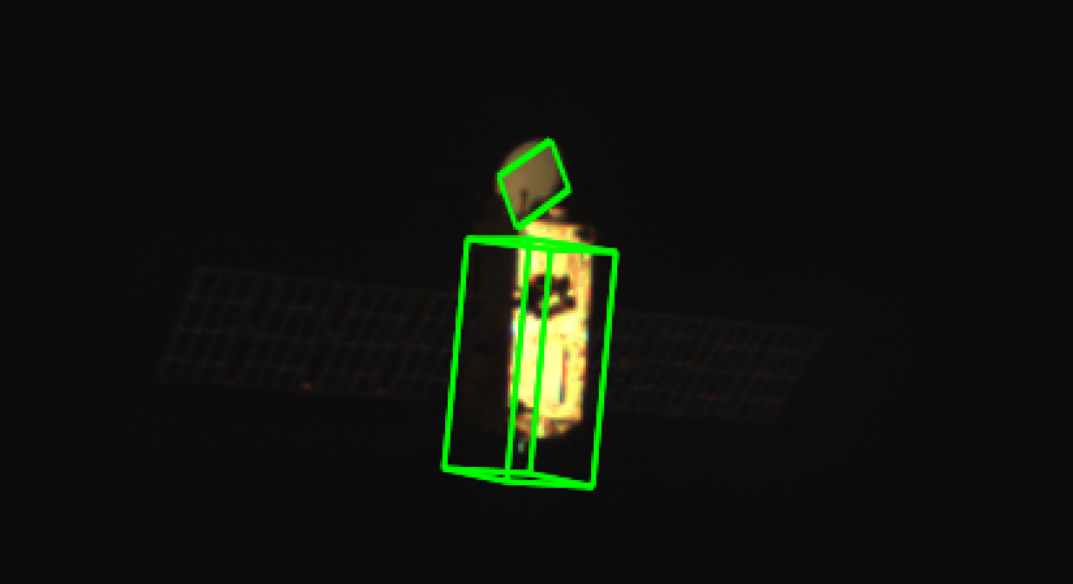}
  	\end{subfigure}\\
  \caption{Qualitative pose estimation performance on frames of the \texttt{CITY/03} laboratory test rendezvous sequences.}
  \label{fig:res-frames-test04-lab}
\end{figure}

\Figref{fig:res-test04} illustrates the evolution in time of the position and attitude estimation errors for the test sequence \texttt{CITY/03}. \Figref{fig:res-frames-test04-lab} qualitatively illustrates these results. It can be observed that the position error is bounded at \SI{35}{\centi\metre} throughout the trajectory, except for the initial transient period. The mean and median error are shown to be approximately half of that, which corresponds to a figure below \SI{6.5}{\percent} of range. The attitude error is kept below \SI{10}{\degree} for the first \SI{85}{\percent} of the sequence, demonstrating that the network is mostly able to separate the translational motion from the rotational one; a degradation of the estimate is observed during the last \SI{4}{\second}, when the target reaches a rotation of \SI{180}{\degree} around the spin axis and the error peaks at about \SI{20}{\degree}, which can be explained by the fact that the training data is biased towards an observation of that specific attitude for larger relative distances. The mean error is approximately \SI{5.5}{\degree} (resp. \SI{3.97}{\degree} median).

%% file: s_conclusion.tex
\section{Conclusion}
\label{sec:conc}

This paper presented ChiNet: a contribution towards deep learning-based, end-to-end, multimodal spacecraft pose estimation for orbital \gls{ncrv}. The proposed method employs a \gls{cnn} as a front-end feature extractor and applies an \gls{lstm}-based \gls{rnn} back-end to model the temporal relationship between incoming frames from an optical camera. Furthermore, \gls{rgb} images are augmented with those captured in the \gls{lwir} band, granting a feature-rich input beyond the visible. The full pipeline is trained according to an innovative multistage optimisation scheme that categorises the learning process in a coarse to fine fashion.

Each of the proposed contributions was individually tested on realistic synthetic data. The addition of the coarse training stage was demonstrated to mitigate spikes in the pose estimation errors originating from sharp reflections of both Earth and sunlight on the solar array. Including the keypoint-based refinement stage improved the average position and attitude errors. The recurrent module eliminated sharp jumps in the estimate of the position, reducing the mean error by half. The inclusion of multimodal \gls{rgbt} image inputs was shown to improve the mean position error in nearly \SI{80}{\percent} and to reduce the mean attitude error in half. 

Overall, ChiNet was shown to generalise well to unseen trajectories, benchmarking a mean range-normalised position error of \SI{2.5}{\percent} per average trajectory and a mean attitude estimation error of \SI{6.9}{\degree} per average trajectory on the sequences of the Astos dataset. The simplest case was shown to be comparable to the classical solution developed in \citep{rondao2021robust}, even surpassing it in terms of position estimation performance. The pipeline required no localisation or segmentation preprocessing to produce an accurate solution. Lastly, the proposed work was benchmarked on experimental data, demonstrating the capability of the network to learn novel situations under a reduced training regime.

Future work might investigate the robustness of the framework towards non-nominal illumination conditions. Another potential avenue to investigate could tackling the problem of domain adaptation in the context of spacecraft pose estimation, whereby a deep network is trained with synthetic images and tested on real data, as the latter are typically scarce prior to the actual mission, but the former can be generated in large quantities.